\documentclass[10pt]{article}
\usepackage[explicit]{titlesec}
\usepackage{hyphenat}
\usepackage{ragged2e}
\RaggedRight

\usepackage{rotating}  
\usepackage{theorem}   
\usepackage{bm}  
\usepackage{amsmath}
\usepackage{booktabs}  
\usepackage{pdfsync}   
\usepackage{graphicx}
\usepackage{latexsym}
\usepackage{amsmath}
\usepackage{fancyvrb}
\usepackage{colortbl}
\usepackage{enumerate}
\usepackage{pifont}
\usepackage{stmaryrd}
\usepackage{textcomp}
\usepackage{paralist}
\usepackage{wrapfig}
\usepackage[table]{xcolor}
\usepackage{lscape}
\usepackage{lipsum}

\usepackage{garamondx}
\usepackage[T1]{fontenc}
\usepackage{amsmath}
\usepackage{graphicx}
\usepackage{xcolor}

\usepackage{geometry}
\usepackage{fancyhdr}
\usepackage[labelfont={footnotesize,bf} , textfont=footnotesize]{caption}

\usepackage{rotating}  
\usepackage{theorem}   
\usepackage{bm}  
\usepackage{amsmath}
\usepackage{booktabs}  
\usepackage{pdfsync}   
\usepackage{adjustbox}
\usepackage{amsmath}
\usepackage{placeins}
\usepackage{afterpage}
\usepackage{multicol}
\usepackage{comment}
\usepackage{floatrow}
\usepackage{lipsum}
\usepackage[utf8]{inputenc}
\usepackage[english]{babel}
\usepackage{csquotes}
\usepackage{rotating}
\usepackage{graphicx}
\usepackage{url}
\usepackage{amsmath}
\usepackage{mathtools}
\usepackage{subfig}
\usepackage{svg}
\usepackage{algpseudocode}
\usepackage{algorithm}
\graphicspath{ {./Figures/} }
\usepackage{longtable}
\usepackage{multirow}

\usepackage[protrusion=true,expansion=true]{microtype}

\usepackage[bitstream-charter]{mathdesign}
\usepackage[T1]{fontenc}
\usepackage{enumitem}
\usepackage{array}

\makeatletter
\renewcommand\tagform@[1]{\maketag@@@ {\ignorespaces {\footnotesize{\textbf{Equation}}} #1.\unskip \@@italiccorr }}
\makeatother
\titleformat{\section}
  {\normalfont}{\thesection}{1em}{\MakeUppercase{\textbf{#1}}}
\titlespacing\section{0pt}{0pt}{-10pt}
\titleformat{\subsection}
  {\normalfont}{\thesubsection}{1em}{\textit{#1}}
\titlespacing\subsection{0pt}{0pt}{-8pt}

\makeatletter
\newcommand\sixteen{\@setfontsize\sixteen{17pt}{6}}
\renewcommand{\maketitle}{\bgroup\setlength{\parindent}{0pt}
\begin{flushleft}
\sixteen\bfseries \@title
\medskip
\end{flushleft}
\textit{\@author}
\egroup}
\makeatother

\usepackage[sort&compress]{natbib}
\makeatletter
\renewcommand\@biblabel[1]{\textbf{#1.}\hfill}
\makeatother

\bibpunct{}{}{,~}{s}{,}{,}
\title{A Novel Active Solution for Two-Dimensional Face Presentation Attack Detection}

\author{
Matineh Pooshideh*$^{a}$ \\ \medskip
$^{a}$Macquarie University, Sydney, Australia \\  \medskip
matineh.pooshideh@mq.edu.au
}

\begin{document}
\justifying
\vspace*{.01 in}
\maketitle
\vspace{.12 in}

\section*{abstract}

    Identity authentication can be identified as the process of verifying one's identity. There are several identity authentication methods, from the range of old username and password systems to the more recent electroencephalography-based authentication systems, among which biometric authentication is of utmost importance.
    This importance is because biometric authentication uses the individual and unique characteristics of the human being to identify their identity, making it very easy to use. Facial recognition is a sort of biometric authentication with various applications, such as unlocking mobile phones and accessing bank accounts. However, presentation attacks pose the greatest threat to facial recognition. A presentation attack is an attempt to present a non-live face, such as a photo, video, mask, and makeup, to the camera. Presentation attack detection is a countermeasure that attempts to identify between a genuine user and a presentation attack. Several industries, such as financial services, healthcare, and education, use biometric authentication services on various devices. This illustrates the significance of presentation attack detection as the verification step.
    In this paper, we study state-of-the-art to cover the challenges and solutions related to presentation attack detection in a single place. We identify and classify different presentation attack types and identify the state-of-the-art methods that could be used to detect each of them. We compare the state-of-the-art literature regarding attack types, evaluation metrics, accuracy, and datasets and discuss research and industry challenges of presentation attack detection.
    Most presentation attack detection approaches rely on extensive data training and quality, making them difficult to implement. We introduce an efficient active presentation attack detection approach that overcomes weaknesses in the existing literature. The proposed approach does not require training data, is CPU-light, can process low-quality images, has been tested with users of various ages and is shown to be user-friendly and highly robust to 2-dimensional presentation attacks.

\section*{keywords}
Presentation Attack, Face Liveness, Face Anti Spoofing

\vspace{.12 in}



\section{introduction}
    This Section begins with a summary of the research and an explanation of the problem statement, which is efficiently detecting two-dimensional presentation attacks. It then offers our contributions and describes how the suggested approach fills the gaps in the literature. The organisation of the paper is outlined at the end.
    
    \subsection{Overview and Problem Statement}
        In recent decades, the Internet has become one of the most important means of communication between individuals. As a result of COVID-19 and the pandemic, various businesses accelerated digitisation~\cite{dorazio2022} and several businesses and service providers use the Internet to interact with their clients. One of the primary keys to confirming the safety of such virtual interactions is trusting in each other's identities. This trust is of utmost importance because various organisations that require high security, such as banking, use the Internet platform to engage with their clients and sometimes perform transactions with enormous value. E.g., according to Forbes~\cite{strohm_2022} 78\% of Americans prefer digital banking. Hence, it is significant for organisations to ensure their clients are who they claim to be.
         
        Network Authentication is a branch of cybersecurity that establishes this trust by verifying users' identities via the Internet. To identify the identity, a three-level identity check could be used: (i) Identity Validation: determining whether the claimed identity exists; (ii) Identity Verification: determining whether the submitted identity information is authentic and genuine; and (iii) Identity Authentication: determining whether the provided information matches the current database information~\cite{experian_2021}. There are numerous types of network authentication systems~\cite{idrus2013review}, such as usernames and passwords, two-factor authentication, and token authentication, among which biometric authentication has gained significant attention in the last two decades.
         
        Biometrics are unique characteristics of each individual, such as the face, fingerprint, iris, and voice. Biometric authentication is the process of identifying individuals' identities through these proprietary features. Due to its ease of use, this field has received significant attention in recent decades~\cite{arab_2022}. Face Recognition (FR) is one of the most popular biometric authentication services, which has gained considerable attention recently. This technology is used in different areas, from the range of phone locks to financial and Know Your Customer (KYC) services~\cite{guennouni2019biometric,ramachandra2017presentation}. In the FR system, similar to all other network authentication services, there is still the need to do the three-level validation, verification, and authentication checks. 
         
        The purpose of the verification check in this system is to determine the genuineness of the face presented to the camera. One type of attack on the FR system is this sensor-based attack, in which a person presents a non-live image of another person's face instead of their own. They might, for instance, use a photograph, video, or a mask of that individual. This type of attack is called Presentation Attack (PA), and the verification check in this system is called Presentation Attack Detection (PAD) or Face Liveness Detection (FLD)~\cite{ramachandra2017presentation}.
    
        Various presentation attack detection approaches have been offered in both academia and industry. In this paper, we propose a novel presentation attack detection approach to detect 2-dimensional presentation attacks. Our approach successfully solves some of the literature gaps by requiring no training data, working with low-quality images, and being CPU-light. We have tested our proposed method against various 2-dimensional attacks, and the results show that our approach is robust to 2-dimensional attacks and user-friendly. In addition, we propose a novel automated online exam framework using some of the identified presentation attack detection methods in the literature.
        
    \subsection{Contributions}
    
        In this paper, we focus on the 2-dimensional presentation attack detection problem. Most presentation attack detection approaches have been developed in academia and industry, utilising deep learning and machine learning techniques. The challenges for these models are:
            (i) Data Collection: Since presentation attack types are diverse, and some of them require access to unique technologies and materials, collecting a sufficient amount of all presentation attack types is highly challenging;
            (ii) Data Quality: The quality of cameras in different applications of presentation attack detection varies widely, and most of the currently existing models require a specific minimum quality to be able to detect the presentation attacks; and
            (iii) Resource Intensity: Most presentation attack detection techniques require significant computational resources.
        This research presents a novel, efficient presentation attack detection approach that addresses the abovementioned difficulties. It does not require training data, is CPU-light, and can process low-quality photos. In addition to resolving these issues, we have tested this service with various users and found it both robust to 2-dimensional presentation attacks and user-friendly.
        
        \begin{itemize}
            \item We propose an end-to-end presentation attack detection process pipeline, including a classification of methodologies and attack types.
            \item We introduce a novel active 2-dimensional presentation attack detection approach that addresses some of the gaps and challenges identified in the literature. Our approach is CPU-light; it works with low-quality camera resolutions and requires no training data.
            \item We offer a new framework for online exam proctoring based on the presentation attack detection methods to enhance the online exam experience for both students and supervisors.
            \item We evaluate our proposed active presentation attack detection approach regarding the system's accuracy and usability and analyse it regarding being CPU-light and working with low-quality frames.
        \end{itemize}

    \subsection{Summary and Outline}
        This Section covered the study's motivation, problem statement, and contributions and provided a brief overview of the topic. The remaining Sections of this paper are organised as follows:
        \begin{itemize}
            \item Section~\ref{chap:2} discusses the background and the related studies on presentation attack detection. It introduces an end-to-end pipeline for PAD approaches and comprehensively discusses the existing presentation attacks. It then explains the state-of-the-art PAD approaches, evaluation metrics, available datasets, challenges, and industry applications regarding the PAD problem. 
            \item Section~\ref{chap:3} introduces our proposed active presentation attack detection approach. It discusses the methodology, preprocessing, active test generation, and liveness evaluation. It also introduces a new framework for online exam proctoring using PAD methods.  
            \item Section~\ref{chap:4} explains the experiments done to prove the claims of our proposed active 2-dimensional presentation attack detection approach and its results. It also evaluates our proposed active PAD test regarding its robustness and usability. 
            \item This study is concluded in Section~\ref{chap:5}, which summarises the proposed approach. Additionally, it provides some future directions to expand the given approach in this work as well as resolve other PAD's challenges and obstacles.
        \end{itemize}


\section{Background and State-of-the-Art}
\label{chap:2}
    
    This Section presents and analyses the related work in the presentation attack detection domain. It first introduces the preliminaries, then gives a detailed introduction to existing presentation attacks, followed by a detailed introduction to the existing state-of-the-art attack detection methods. It also introduces an end-to-end pipeline of presentation attack detection approaches. In the next step, this Section analyses the method selection based on the attack type, evaluation metrics, presentation of attack detection in the industry, datasets, and challenges. It lastly gives a summary of the Section.   
    
    \subsection{Preliminaries}
    \label{section:preliminaries}
        Presentation attack detection is a critical stage in a biometric authentication system. This Section focuses on face PAD, which is the process of determining if the person in the video or photo (the person behind the camera) is a genuine person or a presentation attack. Face PAD systems are also known as face liveness detection or face spoofing detection systems.
        
\begin{figure}[htbp]
            \centerline{\includegraphics[scale=0.18]{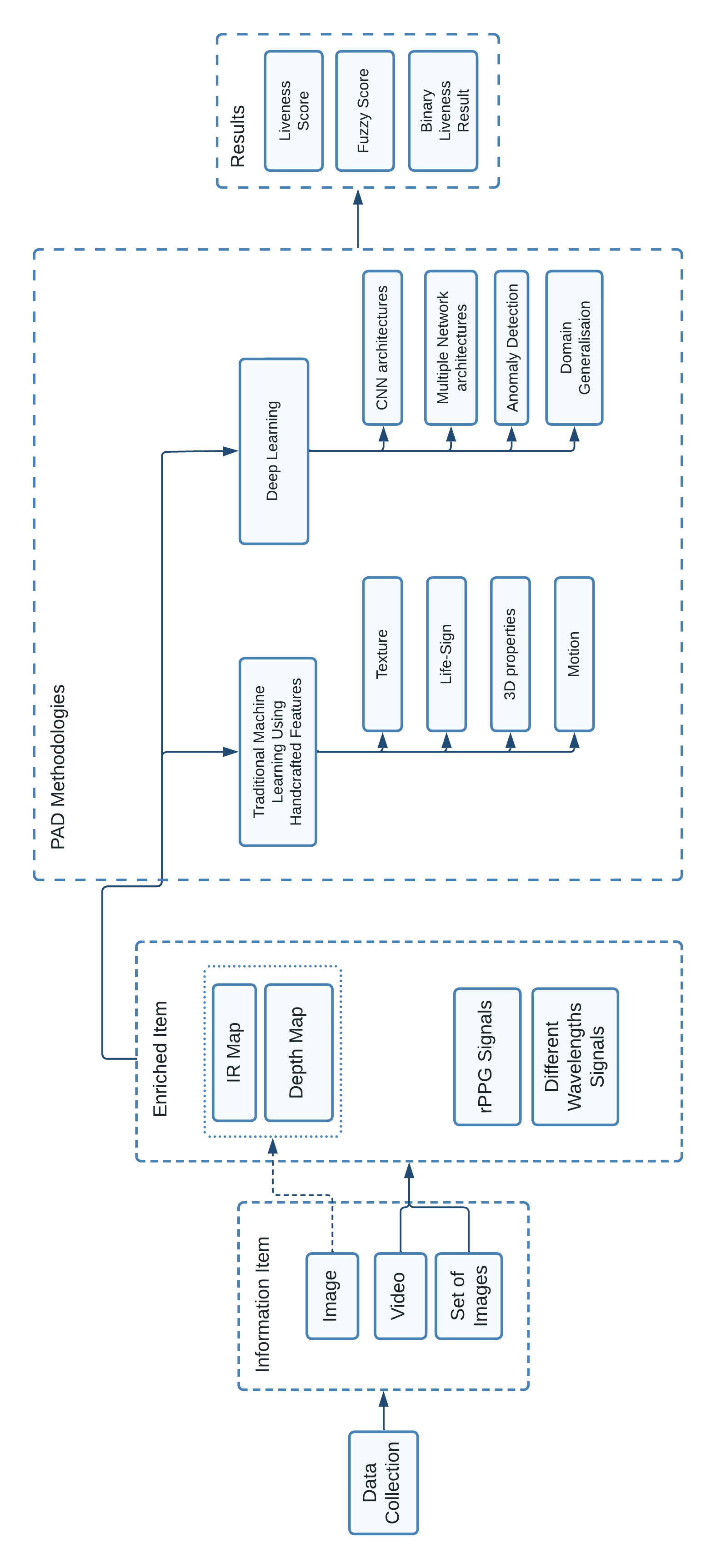}}
            \caption{Presentation Attack Detection Pipeline.}
            \label{fig_taxonomy}
        \end{figure}

        PAD methods are divided into two primary categories: (i) hardware-based: methods that require specific hardware to detect the presentation attacks; and software-based: methods that identify the attacks based on the RGB image/video without requiring any specific hardware~\cite{fatemifar2021client}. In hardware-based methods, some sensors will be used to gather more information from the face and the background. Some methods use specific lights, and others use a specific camera with a specific resolution or the ability to capture various spectral bands, Infra-Red (IR), or other signals and wavelengths from the face~\cite{raghavendra2017extended,sepas2018light}. Other hardware can capture 3-dimensional pictures and depth maps. This Section focuses on software-based methods. The software-based methods' end-to-end pipeline is described in Figure~\ref{fig_taxonomy}.
        
        Software-based PAD methods are divided into two main categories: traditional machine learning (feature engineering using handcrafted features) and deep learning. In the deep learning category, different neural network architectures are used to extract multiple pieces of information from the input to identify if the face on it is live or fake. They are categorised into CNN architectures, multiple network architecture, anomaly detection, and domain generalisation. However, the traditional machine learning or handcrafted features category uses feature engineering to extract face-related features and machine learning algorithms to detect the spoofing cues from the input. The traditional machine learning category is divided into four subcategories: Texture analysis, motion analysis,  Spatial features analysis and life sign analysis. Following is the description of each of these features:
        
        \begin{itemize}
            \item Texture analysis: One of the popular features of a live face, which is different from that of a mask, paper, and media screen, is the texture. This feature is widely used in different papers aimed at detecting 2D and 3D attacks.
            \item Motion analysis: The other feature of a live face vastly used in PAD is motion. The live user can easily move their face in different directions. The facial landmarks would be relatively changed based on the movement's type. Motion analysis can be used in 2D attack detection, mostly used to detect photo attacks.
            \item Spatial features analysis: Also known as 3D properties analysis, is another feature used for PAD. It is widely used in different studies as a part of 2D attack detection. All videos, photos, and in some cases Deepfake videos (only if presented to a camera sensor) are 2-dimensional. 3D features such as the depth map would contribute to recognising 2D presentation attacks.
            \item Life sign analysis: Another helpful feature for PAD analysis is the life sign features. The eye blink, blood flow, mouth movements, and facial expressions are included in this category. Some of them (e.g. blood flow) can be used in 3D attack detection (masks would cover the blood flow), and some of them can be used in detecting 2D attacks (e.g. eye blink detection would detect photo attacks). 
        \end{itemize}
        
        Despite the above categorisation of the PAD methods, two other subsidiary categorisations exist. One way to categorise PAD methods is to divide them based on user interaction. Based on this, the methods are divided into two categories of (i) active methods: (also known as intrusive or interactive methods) in which the user is asked to perform a specific instruction such as moving their face to the sides; and (ii) passive methods: (also known as non-intrusive or non-interactive) in which the user has no interaction with the system. 
        PAD methods are also classified by their input. There are two types of PAD methods based on their input: dynamic and static. Static approaches often employ a single image to extract static information about live faces. On the other hand, dynamic approaches employ multiple frames and videos to detect motion, 3D properties, and dynamic features associated with live faces.
    
    \subsection{Presentation Attacks}
\begin{figure}[htbp]
        \centerline{\includegraphics[scale=.7]{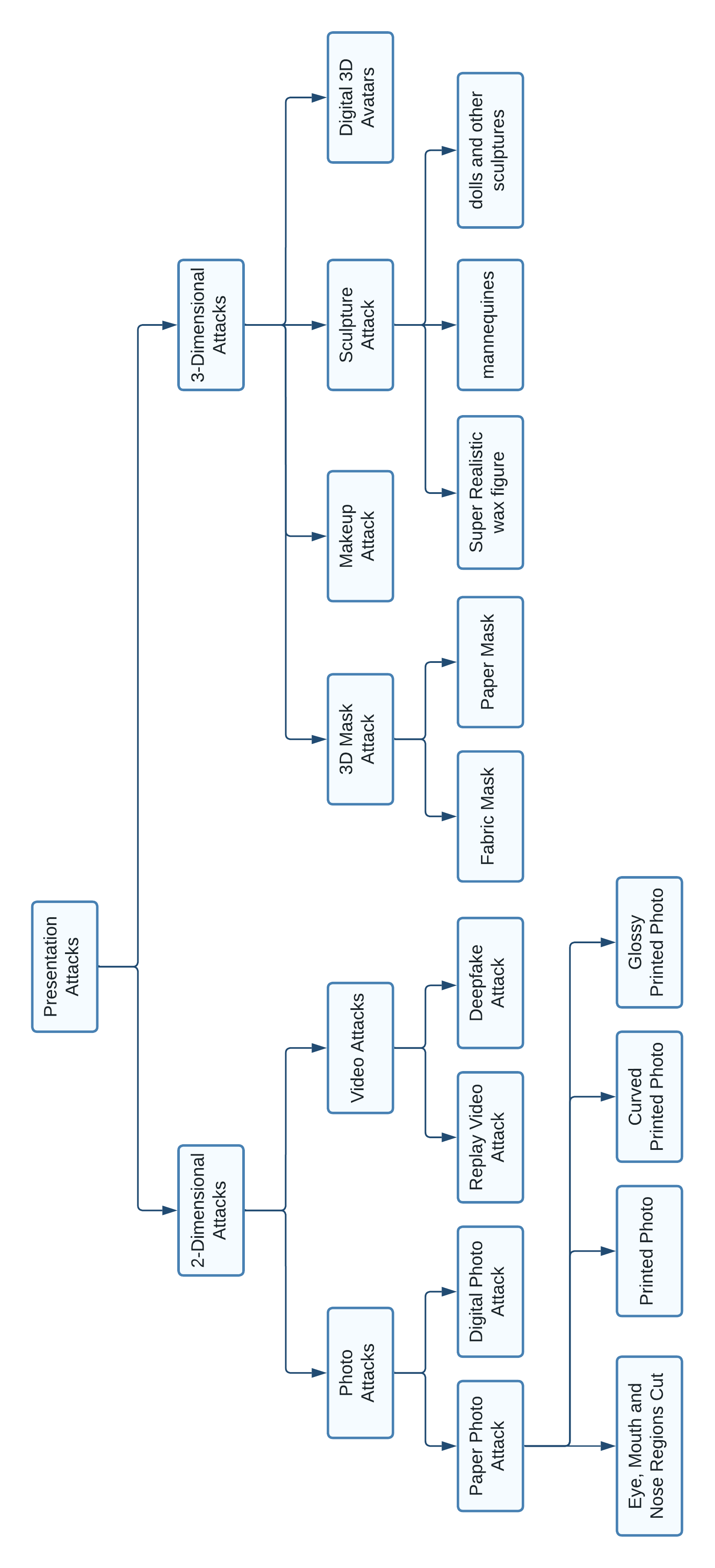}}
        \caption{Diagram of Face Presentation Attack Types.}
        \label{fig: PA diagram}
        \end{figure}

        Scammers attempt to spoof facial recognition systems by displaying a user's fake face to the camera. They utilise various materials to impersonate the user's face, resulting in various presentation attacks. It is critical to understand these attacks and their various kinds to detect them. Knowing the presentation attacks will assist in identifying face-related features that might differ between a presentation attack and a genuine face. These features are then used to identify the attack. Therefore, we comprehensively study the various currently existing presentation attacks in this section. Figure~\ref{fig: PA diagram} shows an overview of all of the studied presentation attack types.
    
\begin{figure}[htbp]
                \centerline{\includegraphics[scale=2.5]{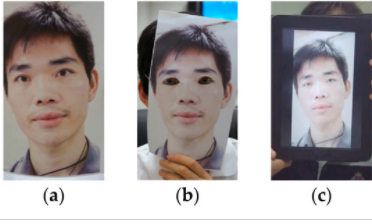}}
                \caption{Examples of 2D attacks: (a) is a print attack; (b) is modified print attack; and (c) is a digital photo attack~\cite{wang2017face}.}
                \label{fig: 2d pa example}
            \end{figure}

        Generally, presentation attacks are classified into two types: (i) 2-Dimensional Presentation Attacks (2D PAs), which refer to attacks in which a scammer uses a plane material with a face to spoof the system, and (ii) 3-Dimensional Presentation Attacks (3D PAs), which refer to attacks in which a scammer uses a 3D material with a face to spoof the system. Examples of 2D and 3D presentation attacks are shown in Figure~\ref{fig: 2d pa example} and~\ref{fig: 3D pa example}, respectively. The details of the two categories are discussed in the following parts.
    
        \subsubsection{2-Dimensional Presentation Attacks}
        \label{section:2d attacks}
        
            \begin{itemize}
                \item Print attack: The face is printed on plane paper. It can be glossy, normal, or any other paper type. 
                \item Digital photo attack: The face is shown on a media screen (mobile phone, PC,  and tablets). This type of attack is classified as different from a print attack due to the different textures. Normally when a camera is capturing frames from a digital screen, extra noise, such as Moiré patterns, can help to detect this type of attack. 
                \item Replay attack: The video of a live face is shown on a media screen. Since this type of attack is a video, we can have eye blinks, facial expressions and different movements. Nevertheless, still, since it is captured from a digital media screen, it can contain different noises.  
                \item Modified print attack: This refers to a print attack in which the paper is cut in some regions or bent. The cut regions are usually the eyes, nose and mouth. The scammer places the cut photo in front of their face to bypass some detection methods, such as eye blink, eye movements, and mouth movements, and to add dimension to the paper by using their nose in the cut region. Another type of these attacks is to bend the paper. This action also adds dimension to the paper.
                \item Deepfake attack: This can be a sub-type of replay video attack if it is presented to the camera sensor. The difference between Deepfake technology and a replay attack is that there is a genuine face in a replay attack, but in Deepfake technology, the face is made artificially. However, if the Deepfake video is not presented to a camera sensor and is being directly sent to a presentation detection system, it might be considered a 3D attack.       
            \end{itemize}
            
        \subsubsection{3-Dimensional Presentation Attacks}
        \label{section:3d attacks}
        
\begin{figure}[htbp]
                \centerline{\includegraphics[scale=1]{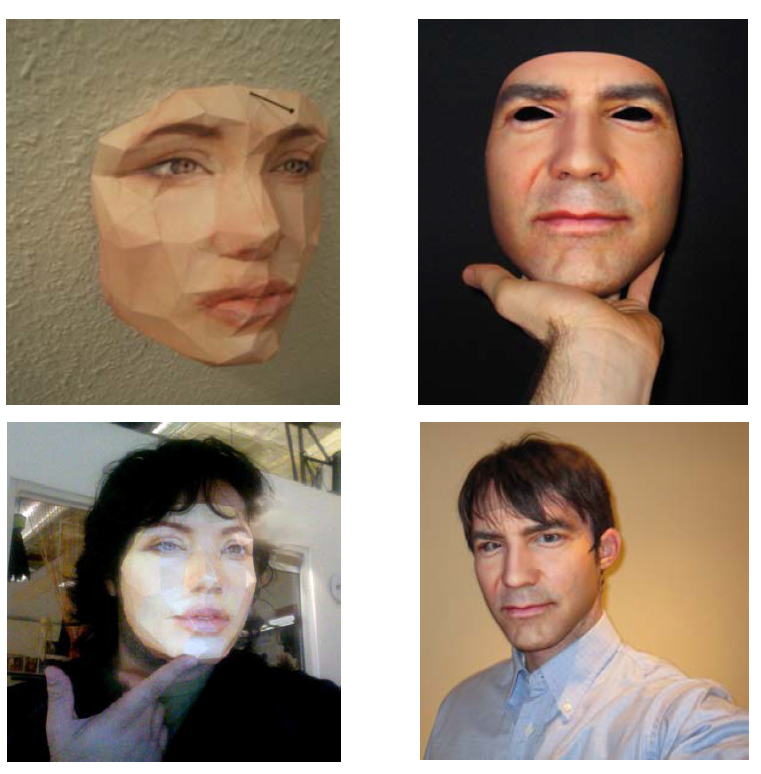}}
                \caption{Example of mask attacks~\cite{kose2013countermeasure}.}
                \label{fig: 3D pa example}
            \end{figure}

\begin{figure}[htbp]
                \centerline{\includegraphics[scale=0.95]{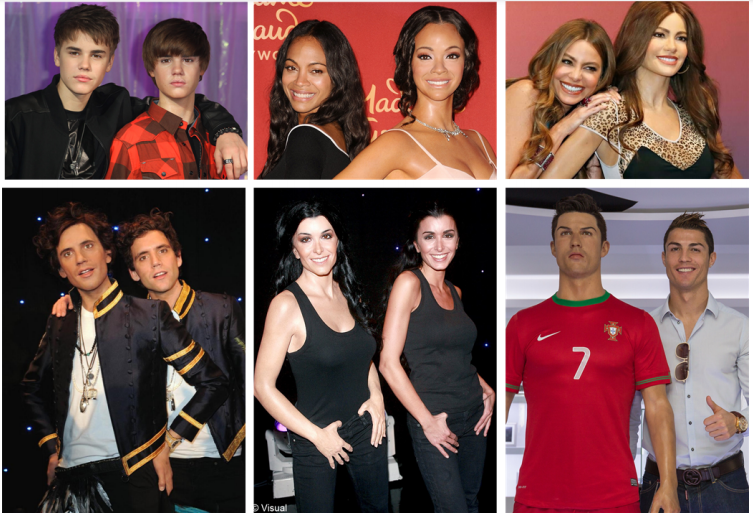}}
                \caption{An example of super-realistic wax figure attack~\cite{jia2019database}.}
                \label{fig5}
            \end{figure}

\begin{figure}[htbp]
                \centerline{\includegraphics[scale=2.5]{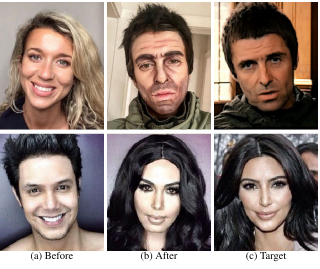}}
                \caption{An example of makeup attacks: (a) Is a scammer; (b) Is the makeup attack; and (c) Is the target user~\cite{rathgeb2020makeup}.}
                \label{fig2}
            \end{figure}

            \begin{itemize}
                \item Mask attack: This is one of the most popular 3D attacks. In this case, the scammer wears a mask of a real person to spoof the system. The quality of masks' materials could range from low to high. E.g., we have paper masks which are not very realistic, or fabric masks, which are of higher quality, as shown in Figure~\ref{fig: 3D pa example}. 
                \item Sculpture attack: For this type of attack, the scammer uses a face sculpture, mannequin, doll, or super-realistic wax figures. An example of this attack is displayed in Figure~\ref{fig5}.
                \item 3D avatar attack: Although it is a face shown on a digital media screen, it does have shadows and dimension simulations that can be categorised as a 3D attack.
                \item Makeup and surgery attack: This refers to the heavy makeup or surgeries that make the scammer's face look like the user's face. Figure \ref{fig2} shows an example of makeup attacks.
            \end{itemize}
             
    \subsection{Presentation Attack Detection}
\begin{figure}[htbp]
            \centering
            \subfloat[]{\includegraphics[width=.5\linewidth]{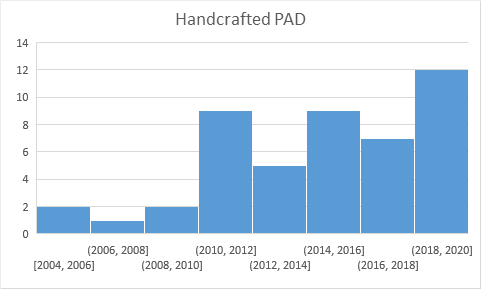}} 
            \subfloat[]{\includegraphics[width=.5\linewidth]{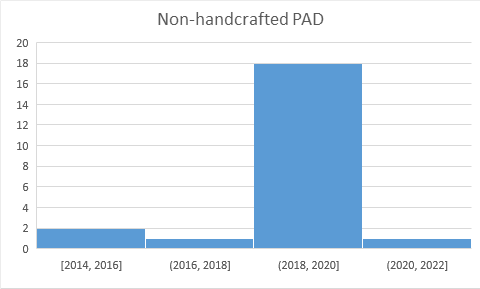}}
            \caption{The histograms of studies using handcrafted and non-handcrafted features in different years.} 
            \label{fig: handcraft vs non} 
        \end{figure}

        In this section, we analyse the PAD methods based on the taxonomy described in the Preliminaries section~\ref{section:preliminaries}. Based on the taxonomy, the PAD methods are generally categorised into handcrafted and non-handcrafted parts. Figure~\ref{fig: handcraft vs non} shows a comparison of histograms of non-handcrafted and handcrafted PAD based on our studies.
        Figure~\ref{fig: handcraft vs non} shows that non-handcrafted features are getting more prevalent in recent years, and they started to gain attention around 2013. Nevertheless, handcrafted methods are older, have existed since 2004 and are still vastly used in different PAD studies. The detailed analysis of each of the two main categories is discussed in the following subsections.
            
        \subsubsection{Handcrafted Feature}
        \label{section:handcraft vs DL}
                
\begin{figure}[htbp]
                \centerline{\includegraphics[scale=0.7]{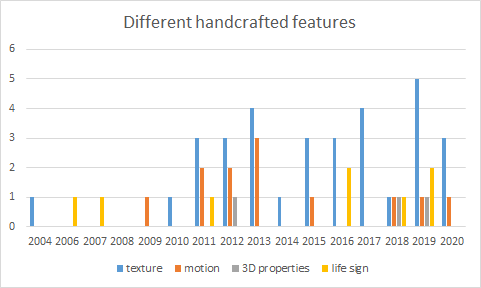}}
                \caption{The number of studies using different handcrafted features in different years.}
                \label{fig:handcraft based}
            \end{figure}

            Handcrafted methods are divided into four different types based on taxonomy. Texture analysis has been drastically used in different years, and it gets more popular each year. This feature proved to be more robust and popular based on our studies. Motion analysis is the second most popular method after texture analysis. However, in recent years life-sign is also getting popular. 3D properties have been used sparingly in handcrafted methods. It is, however, one of the methods that have been used a lot in hardware-based approaches since, with infrared cameras and other hardware, it is easy to get the depth information of the face. Figure~\ref{fig:handcraft based} shows the chart related to the number of different handcrafted features in different years used in our studies. 
                
            \par{Texture Analysis}
            \label{section:texture}
            
\begin{figure}[htbp]
                    \centerline{\includegraphics[scale=0.65]{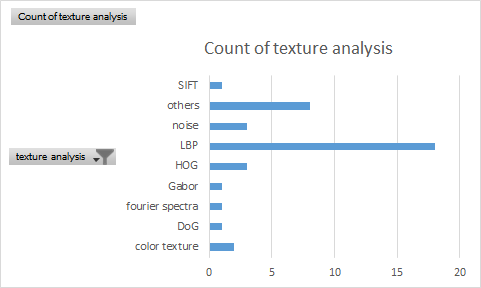}}
                    \caption{The number of studies using different texture analysis methods.}
                    \label{fig10}
                \end{figure}

                Texture analysis is the most popular feature in handcrafted methods and has been used widely in different years and studies. Figure \ref{fig10} shows different texture analysis methods. The Local Binary Patterns (LBP) method has gained the most attention among various texture analysis methods. More than 45 per cent of studies in handcrafted methods have used LBP as their texture descriptors. 
                
\begin{figure}[htbp]
                    \centerline{\includegraphics[scale=0.5]{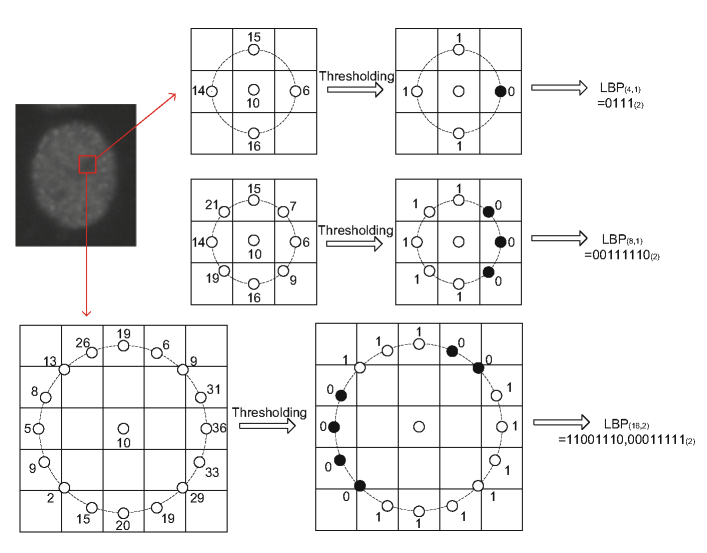}}
                    \caption{The calculation of local binary patterns with different radius and neighbours~\cite{ebrary}.}
                    \label{fig11}
                \end{figure}

                LBP works in the following way; For each pixel, it uses n neighbours of that pixel in the radius of r. Then it compares the pixel with each of those neighbours. If the pixel's value is greater than its neighbour, then the neighbour gets 0; otherwise, the neighbour gets 1. Then clockwise or counterclockwise, it reads the binary numbers that neighbours have made. Then it converts that number to a decimal number and adds it to the corresponding pixel's cell in the result array. Figure \ref{fig11} shows the calculation of LBP.
                
                Calculating local binary patterns with different radiuses and neighbours can create different textures. The histogram of different LBP calculations is used in different papers as the feature vector for PAD. Maatta et al.~\cite{maatta2011face} use the histogram of three different calculations of LBP. One with 16 neighbours and radius 2, one with eight neighbours and radius two and one with eight neighbours and radius 1. For the LBP with eight neighbours and radius 1, the image is divided into smaller blocks, and the histograms of each block are concatenated. In the end, all the histograms are concatenated and fed to an SVM to classify the image as fake or real. They train their model on 2D attacks. Later, Kose et al.~\cite{kose2013countermeasure} use the same method and trains it on 3D attacks to calculate the effectiveness of this method on 3D attacks. 
                
                In another study, Maatta et al.~\cite{maatta2012face} use the histogram of LBP and two other feature descriptors, Gabor wavelets and Histogram of Oriented Gradient (HOG), with a score-level fusion model. Chingovska et al.~\cite{chingovska2012effectiveness} apply various LBP calculations and analyse their effectiveness on PAD. Freitas Pereira et al.~\cite{freitas2012lbp,freitas2014face} use a dynamic approach for LBP, Local Binary Pattern from Three Orthogonal Plane (LBP-TOP), which uses a sequence of images and calculates the LBP for XY, XT, and YT planes. In another study, they use frame correlation, LBP and LBP-TOP to analyse their effectiveness in real-world scenarios using inter-dataset and intra-dataset testing~\cite{de2013can}. Boulkenafet et al.~\cite{boulkenafet2015face} apply LBP on different colour channels, Y, Cb, and Cr, and concatenate their histograms to create the feature vector. Several studies also use different LBP settings either as the baseline or one of the feature extractors in their PAD methods and analyses~\cite{arashloo2017anomaly,bharadwaj2013computationally,chen2019cascade,fatemifar2021client,komulainen2013complementary,liu20163dd,patel2015live,peng2018face,peng2020face,rehman2020enhancing,wang2017face,zhang2020face}. 
                
                Besides LBP, other texture descriptors are used in PAD, such as Scale-Invariant Feature Transform (SIFT), HOG, colour descriptors, and Gabor. They extract different textural patterns in the image and are used to differentiate fake and live attempts in PAD. Some studies~\cite{fatemifar2021client,maatta2012face,rehman2019perturbing} use HOG as one of their features, Patel et al.~\cite{patel2015live} use SIFT and LBP as their feature descriptors, and Gabor is used by Maatta et al.~\cite{maatta2012face}. Some others use colour descriptors as one of their feature vectors\cite{peng2020face,wang2017face}. 
                
                Several studies~\cite{ma2020multi,nguyen2019face,pinto2015face,yan2012face} use different techniques to detect noises in the images or videos for presentation attack analysis.  Fourier transformation to detect the focus of the images was one of the earliest methods used in PAD. Li et al.~\cite{li2004live} use Fourier transform as a texture analysis method to detect attacks. Another method used in texture analysis is the Difference of Gaussians (DoG).  DoG subtracts two gaussian blur versions of a photo. Peixoto et al.~\cite{peixoto2011face}, and Tan et al.~\cite{tan2010face} use DoG and Variational Retinex-based Method for PAD. 
                
                Other methods are also used in the texture analysis category. Peixoto et al.~\cite{peixoto2011face}, and Tan et al.~\cite{tan2010face} use the Variational Retinex-based Method, Agarwal et al~\cite{agarwal2016face} use Haralick features, Tronci et al.~\cite{tronci2011fusion} use different feature descriptors, Li et al.~\cite{li2016face} and Van et al.~\cite{van2019face} use shearlets, and Patel et al.~\cite{patel2016secure} use image distortion for PAD.
                
            \par{Life Sign Analysis}
            \label{section:Life-Sign Analysis}
                
                Life sign analysis is a method that detects some features that are a part of a live face's natural behaviours, such as different facial expressions and eye blinks. Figure~\ref{fig:handcraft based} shows that life sign was used from around 2006 to 2011 and could have gained little attention at the time. Then, around 2016 it gradually gained attention again. The reason could be the advent of using the more powerful life-sign methods. They detect face pulsed or blood flow on the skin, help the systems detect more attack types, and it was gradually shown that they have pretty good accuracy and are reliable.
                
                Several methods are used to detect the eye blink as a life sign~\cite{jee2006liveness,pan2007eyeblink,tronci2011fusion}. These methods were able to detect simple photo attacks. Jee et al.~\cite{jee2006liveness} use a binary threshold to binarise the eye region in different frames and compare them with hamming distance to detect whether they are getting enough movements. CRF method is also used to find blink sequences in different studies~\cite{pan2007eyeblink,tronci2011fusion}. 
                
\begin{figure}[t]
                \centerline{\includegraphics[scale=0.5]{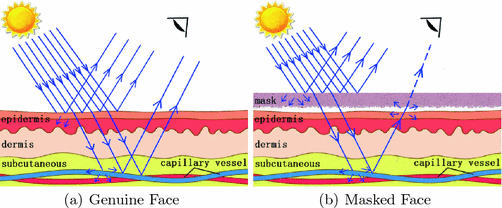}}
                \caption{The light reflection of the real skin with blood flow and the fake attempt~\cite{liu20163d}.}
                \label{fig12}
                \end{figure}

                As mentioned earlier, another method used in the life-sign category is a method that extracts the blood flow or pulses. Remote photoplethysmography (rPPG) signal is the method used in this regard. It is shown that we can extract the heart rate from the RGB channels of the video of the skin. The light reflected from the skin in different frames is different due to the heart rate, as shown in Figure \ref{fig12}, and this difference can be extracted from the RGB channels of frames. Several studies use rPPG signals to detect various types of attacks~\cite{li2016generalized,lin2019face,liu20163d,liu2018learning}. 
                
            \par{Spatial Features Analysis}
            \label{section:spatial}
            
\begin{figure}[h]
                    \centerline{\includegraphics[scale=0.25]{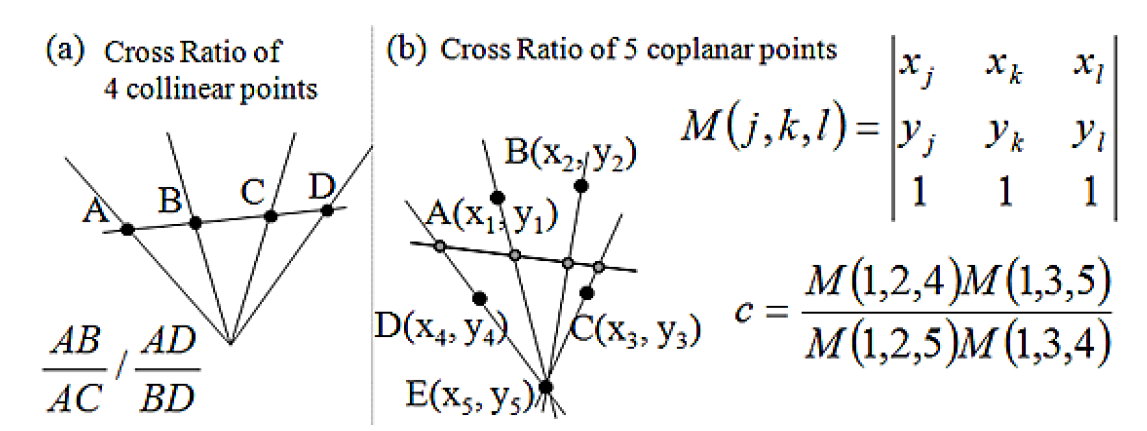}}
                    \caption{The calculation of cross ratios of five co-planar points(b) and four co-linear points(a)~\cite{de2012moving}.}
                    \label{fig13}
                \end{figure}

\begin{figure}[t]
                    \centerline{\includegraphics[scale=0.25]{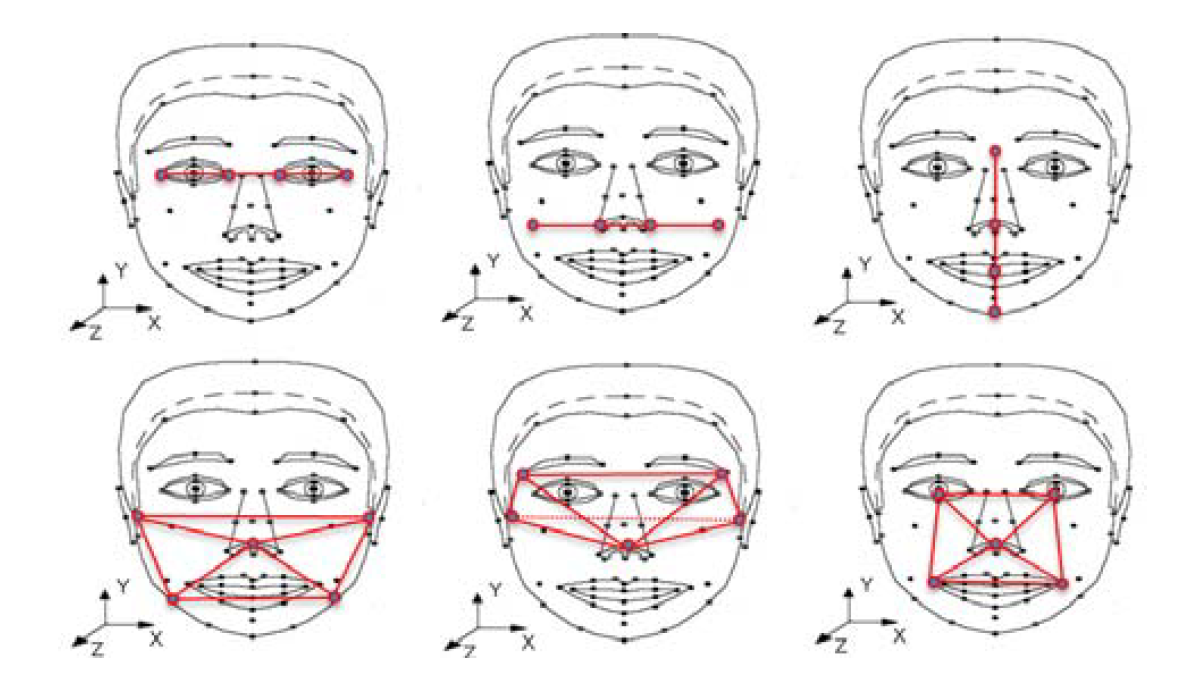}}
                    \caption{The co-linear and co-planar points chosen by Marsico et al.~\cite{de2012moving}.}
                    \label{fig14}
                \end{figure}

                Spatial features are primarily used in hardware-based methods since they are able to extract robust depth information from the face. In our studies, most methods that extract depth maps or depth information about the face are in the non-handcrafted category, and they extract depth maps using deep learning approaches. However, one paper has used a mathematical lemma to detect the 3D properties of the face. Marsico et al.~\cite{de2012moving} use a mathematical lemma that claims if we have five points with a specific situation in one plane (five co-planar points), then a cross-ratio exists that is rotation invariant. Also, if there are four points in a single line in a single plane (four co-linear points), then a cross-ratio exists that is rotation invariant. See Figure~\ref{fig13} for more detail. Marsico et al.\cite{de2012moving} use 6 different sets of points on the face which are not in the same plane and line when the face is in 3D but are in the same plane and live if the face is a photo attack. Finally, by asking the user to move to the sides, this method is able to detect photo attacks. Figure~\ref{fig14} shows the points this paper uses. 
                
            \par{Motion Analysis}
            \label{section:motion}
            
\begin{figure}[htbp]
                    \centerline{\includegraphics[scale=0.7]{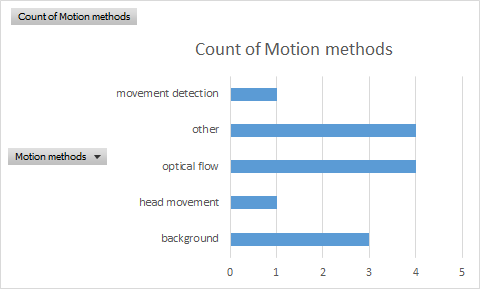}}
                    \caption{The number of studies using different motion analysis features.}
                    \label{fig15}
                \end{figure}

                Motion analysis is another handcrafted feature used in presentation attack detection. Our studies show that after Texture, motion analysis has been used most; Figure~\ref{fig10}. The motion methods are all dynamic and need multiple frames, and the aim is to follow specific movements in the facial area or the background. They are used for detecting 2D attacks. Figure~\ref{fig15} shows the related methods used in motion analysis. 
                
                Optical flow is a method to track a point in different frames. The direction of the movement and its velocity can be extracted using optical flow. This method has widely been used in different studies~\cite{bharadwaj2013computationally,kim2011motion,kollreider2009non,smiatacz2012liveness}. E.g., Kollreider et al.~\cite{kollreider2009non} use two areas of the face (ears and nose) and track the motion of those areas using optical flow. The idea of tracking the motion of these areas is that in a 3D-shaped face when the face is turned in a direction, the nose will move more than the ear since it is closer to the camera.
                Furthermore, the ear moves in the opposite direction of the nose. Through this analysis, this study is able to detect photo attacks while requiring the user to interact with the system and is an intrusive test. 
                
                Another method used in motion analysis is the use of the background. This is by tracking different frames and seeing whether the background is moving in the same direction as the face and with the same velocity~\cite{fourati2020anti,yan2012face}. Another way of tracking background is to compare the background before the face is shown to the screen and the background after the face is shown to the screen. This will help to understand if the background has changed if a video or photo attack is detected~\cite{kim2011motion}. 
                
                The other method that is used in motion tracking is movement detection. It is a simple method of frame-by-frame calculating the ratio of pixels that have moved and the total number of pixels~\cite{tronci2011fusion}. Head movement is another way of detecting whether the moving object is a printed photo or not~\cite{lavrentyeva2018interactive}. There are some other methods used in motion analysis as well, such as frame correlation~\cite{de2013can,komulainen2013complementary}, motion patterns~\cite{ma2020multi} and motion magnification~\cite{bharadwaj2013computationally,pinto2015face}.

        \subsubsection{Non-handcrafted Features}
        \label{section:non-handcrafted features}
            
            \par{Deep Learning}
            \label{section:dl}
            
                Different deep-learning approaches have been used for presentation attack detection. Since we are using images, most of the deep learning approaches use new or pre-trained CNN structures~\cite{akbulut2017deep,arashloo2020unseen,baweja2020anomaly,bousnina2021unraveling,chen2019cascade,fatemifar2021client,george2020learning,jia20203d,koppikar2021face,koshy2019optimizing,lavrentyeva2018interactive,li2020compactnet,lin2019face,rehman2019face,lin2019convolutional,liu2018learning,liu2019presentation,ma2020novel,mu2019face,patel2016cross,rehman2020enhancing,rehman2019perturbing,sun2020face,tu2020learning,wu2019face,yang2014learn,yu2019diffusion,zhang2020face}. Some of these studies use deep learning to extract features and then feed them to a classifier or create a fusion model with other handcrafted features~\cite{chen2019cascade,fatemifar2021client,jia20203d,lavrentyeva2018interactive,lin2019face,liu2018learning,liu2019presentation,rehman2020enhancing,rehman2019perturbing} and some others are end to end neural networks~\cite{khatami2020convolutional}. Another structure that has been used is the Recurrent Neural Network (RNN) structure. RNNs are used as a tool to detect rPPG signals and are a part of a bigger network~\cite{liu2018learning}. Some other studies use attention networks to extract attention-aware features from the faces~\cite{jia2021face,wu2019face}. 
                
            \par{Anomaly Detection and Domain Generalization}
            \label{section: anomaly detection}
            
                Most studies use binary classifiers or binary classification neural networks to detect whether the input is fake or genuine. Hence they use both real and fake samples in the training stage, and the systems are only robust to previously known attacks. The problem is that the variety of attacks is growing as technology grows. Also, various types of attacks already exist, and it is challenging to have a database containing all these types. Therefore, it is better to have a system that can also be robust to unseen attacks. In order to solve this issue, in the last five years, several studies have been done on the advantages of using anomaly detection (one-class classifiers) and domain generalisation methods for detecting the liveness. 
    
                Anomaly detection methods detect outliers and rare cases that differ from the standard samples in a model. In the training stage, these systems only use actual samples and can detect unseen attacks as anomalies at the operation and test levels. Anomaly detection considers real faces as standard samples and all types of attacks as anomalies in the PAD problem. In handcrafted features, one-class classifiers are used instead of binary classifiers. In neural networks, they are designed so that they only contain real images in the training stage.
                
                Arashloo et al.~\cite{arashloo2020unseen} take the first step in this direction and use anomaly detection to detect unseen attacks. They use three handcrafted feature extractors to extract features from video sequences. They also use Support Vector Data Description (SVDD) and One-Class Sparse Representation-based Classification (OC-SRC)  for anomaly detection. Arashloo et al.~\cite{arashloo2020unseen} train each feature vector with different binary classifiers and one-class classifiers and compare all models to evaluate the effectiveness of using one-class over binary classifiers. The results of the evaluation part show that both one-class and two-class classifiers are operating almost the same and could be better in confronting unseen attacks. More studies are needed in this area.
                 
                Following this research, Nikisins et al.~\cite{liu2019presentation} conducted a similar study. They use some handcrafted features as feature vectors. They train the feature vector on logistic regression and a Two-Class Support Vector Machine (TC-SVM) for binary classification baseline. For the anomaly detection method, they train feature vectors on one-class SVM and	One-Class Gaussian Mixture Model (OC-GMM). The results show that two-class classifications outperform anomaly detection methods when attacks are seen in the training stage. However, two-class classifications do not operate well and are stable regarding unseen attacks, and anomaly detection methods, especially one-class GMM, outperform the binary classifiers. 
                
                Besides handcrafted features and one-class classifiers, some neural network anomaly approaches have also been studied in recent years. Perez-Cabo et al.~\cite{perez2019deep} use a deep metric learning approach for the PAD. Metric learning approaches try to maximise inter-class separability and minimise intra-class variance. This approach will compact the real class and separate it from other fake classes as much as possible. Perez-Cabo et al.~\cite{perez2019deep} utilise a triplet loss function to construct a metric learning approach using three images together as input representing anchor, positive and negative. The triplet loss function tries to minimise the anchor-positive pair's variance and maximise the anchor-negative pair's distance. To follow an anomaly detection approach, they define two categories of closed-set and open-set addressing classes that can be modelled entirely in the training stage and cannot be entirely modelled in the training stage, respectively. This anomaly detection approach focuses on building a discriminative embedding function rather than classification, and the authors propose a novel softmax function for this aim. At the operation level, this system gets an image and returns a feature vector in the embedding space. In order to get the genuine class probability, an SVM is used to learn the boundaries of the genuine class and decide whether the feature vector is inside the genuine class or not. The final results show that this deep anomaly detection approach outperforms the state-of-the-art binary classification methods.
                
                Another study has used a deep anomaly detection network architecture as an end-to-end system. Baweja et al.~\cite{baweja2020anomaly} propose a network architecture using the Gaussian model and VGGFace as a pre-trained CNN. The idea is taken from OC-SVM and defines a hyperplane that can distinguish real and fake data. To this aim, a batch of genuine samples is given to a pre-trained model (VGGFacee), and the output features are concatenated to define a feature vector. Next, pseudo-negative features are created based on a Gaussian distribution with a mean calculated based on the mean of the current batch and the previous batch's mean. After that, the feature vector of genuine samples and that of the pseudo-fake samples are concatenated and fed to a one-class classifier which detects the samples' class with a cross-entropy loss function. This approach is then compared with different anomaly detection approaches such as OC-SVM, SVDD, Mahalanobis Distance (MD), OC-GMM and one-class-CNN (OC-CNN, which is the same method used by Baweja et al.~\cite{baweja2020anomaly} except that the mean of the Gaussian model in OC-CNN is always zero). It is indicated that their method outperforms the other mentioned anomaly detection baselines.
                
    \subsection{Method Selection Based on the Attack Type}
    \label{section6:effects of pad}
        
        Based on our primary taxonomy, each type of method could be used for some particular types of attacks. For example, some methods only cover photo attacks, some could cover 2D attacks, and some methods can cover 3D attacks. This section discusses the types of attacks that each of the taxonomy methods can cover. Spatial features are only used for 2D attack detection, which is evident since the components extracted in this taxonomy all relate to the 3D properties of the face.
        
        Motion properties are highly related to the Three-dimensional properties of the face and detecting moving backgrounds of the fake 2D attacks. Hence, this type of method covers only 2D attacks. A 3D attack can easily have movements like the genuine face and cannot be covered by motion detection methods. In our research, different studies have used motion analysis. For example, some studies have used motion techniques that can only cover photos~\cite{kollreider2009non,lavrentyeva2018interactive,smiatacz2012liveness,tronci2011fusion}, and some others have used motion detection methods to cover all types of 2D attacks~\cite{bharadwaj2013computationally,de2013can,kim2011motion,komulainen2013complementary}. Some studies that use motion are able to cover 2D and 3D attacks~\cite{fourati2020anti,pinto2015face}; However, they have used multiple methods, and the motion detection part in these papers is also used only for 2D attack detection.
        
        Life sign analysis can be used to detect 2D and 3D attacks. Different life sign properties can be used for detecting various types of attacks. E. g., eye blinking can happen in videos, photos or masks with the eye region being cut, so there are better choices to detect liveness in these scenarios. However, sculptures or photos cannot have an eye blink, making this test robust in detecting these attacks. Nevertheless, pulse detection or blood flow detection can help detect different surfaces, such as masks, videos on digital screens and photos on paper or other tools. Hence it can cover all attack types. In our studies, several methods are focusing on life-sign, which, based on the method, can cover photo attacks~\cite{jee2006liveness,pan2007eyeblink,tronci2011fusion}, mask attacks~\cite{liu20163d}, and both 2D and 3D attacks~\cite{li2016generalized,lin2019face}.
        
        Texture methods can be used to detect both 2D and 3D attacks. Like life signs, texture analysis can be used to detect different attack types based on the technique. Some texture descriptors are more detailed and robust and can be used in high-quality 3D as well as 2D attacks. In our studies, texture analysis is used primarily for photo attack detection~\cite{li2004live,maatta2011face,maatta2012face,nguyen2019face,peixoto2011face,tan2010face,tronci2011fusion,yan2012face} and 2D attack detection~\cite{arashloo2017anomaly,bharadwaj2013computationally,boulkenafet2015face,chingovska2012effectiveness,de2013can,freitas2012lbp,freitas2014face,komulainen2013complementary,li2016face,ma2020multi,nowara2017ppgsecure,patel2016secure,patel2015live,peng2018face,peng2020face,wang2017face}. In some studies, texture analysis is used for both 2D and 3D attack detection~\cite{agarwal2016face,pinto2015face,van2019face}. 
        
        Non-handcrafted features use different deep-learning structures to detect the attacks. Therefore, they are similar to the texture and life sign analysis and can detect 2D and 3D attacks based on the various structures and information it is extracting. In our studies, most of the non-handcrafted methods are used for 2D attack detection~\cite{akbulut2017deep,arashloo2020unseen,chen2019cascade,koppikar2021face,koshy2019optimizing,lavrentyeva2018interactive,lim2020one,lin2019convolutional,liu2018learning,liu2019presentation,ma2020novel,mu2019face,patel2016cross,rehman2020enhancing,rehman2019perturbing,rehman2019face,sun2020face,tu2020learning,wu2019face,yang2014learn,yu2019diffusion,zhang2020face,zhou2020domain}. In some studies, they are used for all types of attacks~\cite{baweja2020anomaly,bousnina2021unraveling,fatemifar2021client,george2020learning,jia2021face,jia20203d,li2020compactnet,lin2019face}. Although texture analysis and deep learning techniques of different studies are mainly used for 2D attack detection, they can be tested on 3D attacks. Furthermore, many methods have been created when no proper 3D attack dataset existed. Thus, the results can be evaluated for 3D attacks and future work and see if they can be used accurately.
    
    \subsection{Evaluation Metrics}
    \label{section: evaluation metrics}
\begin{figure}[htbp]
                \centerline{\includegraphics[width=0.6\linewidth]{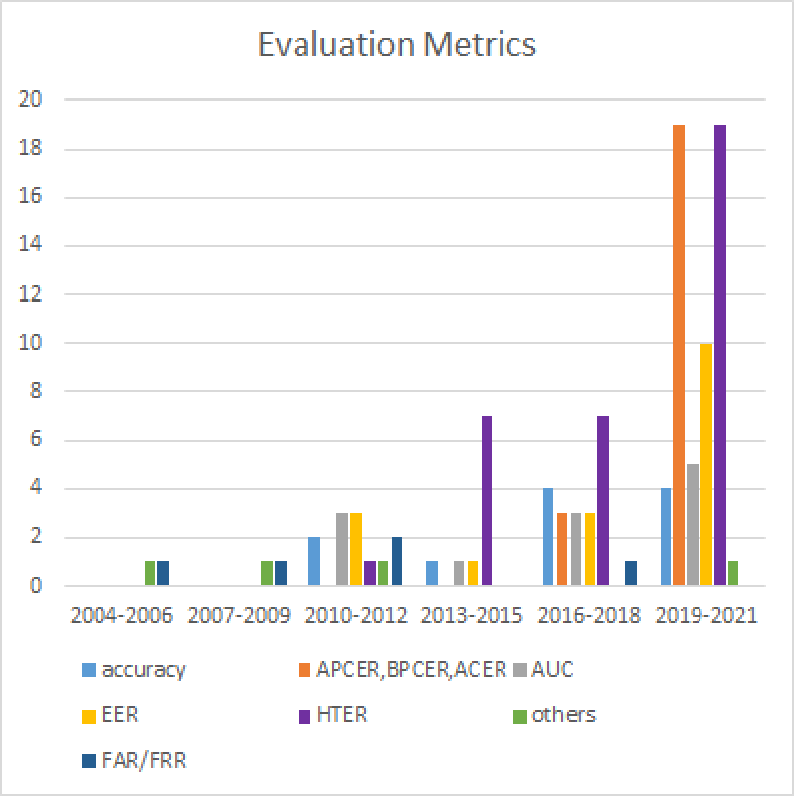}}
                \caption{The frequency of using different evaluation metrics during the last two decades.}
                \label{fig:evaluation}
            \end{figure}

            The evaluation of a presentation attack detection method generally addresses the system's robustness against presentation attacks as well as its ability to detect genuine users reliably. Various evaluation metrics for PAD approaches have been utilised during the last two decades. Figure~\ref{fig:evaluation} depicts the various metrics utilised in different years.
            
            In early studies, the metrics False Acceptance Rate (FAR), also known as False Positive Rate (FPR), and False Rejection Rate (FRR), also known as False Negative Rate (FNR), were used. FAR is the error rate when a fake case is identified as real, while FRR is the error rate when a real instance is recognised as fake. PAD methods can be seen as binary classifiers with two classes: fake and real. The main thing in PAD systems is that no fake should be recognised as real. Therefore, it is critical to keep the FAR as low as feasible while keeping the FRR relatively low.  
            
\begin{figure}[t]
                \centerline{\includegraphics[scale=0.5]{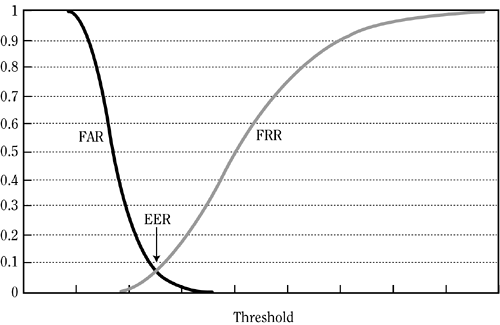}}
                \caption{The EER which is where the difference between FAR and FRR is minimum~\cite{alahmid_2020}.}
                \label{fig: eer}
            \end{figure}

            As shown in Figure~\ref{fig:evaluation}, since 2012, the Half Total Error Rate (HTER), which is the average of FAR and FRR, has been commonly used in PAD approaches. Another popular evaluation metric is Equal Error Rate (EER), where the difference between FAR and FRR is minimal. Figure~\ref{fig: eer} depicts the EER concerning FAR and FRR. Some studies have used Area Under The Curve (AUC), Receiver Operating Characteristics (ROC) curve, and Accuracy. However, they do not describe system performance as well as other metrics.
              
            In 2016, an ISO Standard~\cite{iso_2021} for the Biometric Presentation Attack Detection evaluation metrics was defined. The terms Attack Presentation Classification Error Rate (APCER), Bona-fide Presentation Classification Error Rate (BPCER) and Average Classification Error Rate (ACER) were defined and corresponded to FAR, FRR and HTER, respectively. Therefore, most articles have employed these three standard evaluation metrics for PAD approaches since then. The formulas for the most commonly used metrics are as follows~\cite{chingovska2019evaluation}:\\
            
            \begin{equation}
            \label{eqn:1}
                FAR = APCER = \frac{\# False Positive Samples}{\# Negative Samples}
            \end{equation} 
        
            \begin{equation}
            \label{eqn:2}
                FRR = BPCER = \frac{\# False Negative Samples}{\# Positive Samples}
            \end{equation} 
        
            \begin{equation}
            \label{eqn:3}
                HTER = ACER = \frac{(FAR + FRR)}{2}
            \end{equation}

    \subsection{Presentation Attack Detection in Industry}
        Presentation attack detection is one of the verification checks needed for face recognition systems, and it is essential to have an antispoofing check whenever face recognition is used. Face recognition and all biometric authentication systems have various use cases in different industries. This section will discuss some critical use cases of face recognition and liveness detection in different sectors. Also, we will introduce some of the industry-level PAD approaches.
        
        \subsubsection{Industry Applications}
            In this section, we will discuss the industrial applications of biometric authentication systems and specifically focus on face recognition applications. Presentation attack detection would also be the verification part of the FR systems of all those applications. The following summarises nine different industrial use cases of face recognition systems as the biometric authentication service~\cite{thalesgroup, guennouni2019biometric, hernandez2021biometric, gavin_2021, cordennebrewster,benatallah2016process,beheshti2016scalable,beheshti2018iprocess}.
        
            \begin{itemize}
                \item Law Enforcement: Facial Recognition and other biometric authentication systems are used in law enforcement agencies for criminal investigations. Live Face Recognition is also used for real-time identification of faces in crowds at borders, airports, cities and other public places~\cite{schiliro2018icop,beheshti2020intelligent,schiliro2020novel,schiliro2020cognitive}.
                \item Border Control and Travel: In border control, biometric passports are used as the second generation of electronic passports (e-passports) containing two fingerprints and one photo. The face recognition system automates and speeds up the check-in process in airports, airlines, and hotels. 
                \item Healthcare: Biometric authentication in health care is used by doctors to access patients' medical records using their faces quickly. Also, face recognition is used in caregiving centres to keep track of patients~\cite{ghodratnama2021summary2vec}.
                \item Population and Voter Registration: Face Recognition can help perform the "one person, one vote" law in a vote or population registration~\cite{beheshti2019datasynapse}.  
                \item Access Control: One of the most common uses of facial recognition systems, is to give access to the right people. This access is sometimes physical such as building door or car access, and sometimes the access is logical, such as the phone unlock. In logical access control, face recognition and generally biometric authentication systems replace old username-password systems~\cite{beheshti2017coredb,beheshti2018corekg}.
                \item Commercial: Recently, in most businesses and companies, customer onboarding and account opening have been done online. Know Your Customer (KYC) applications use biometric authentication systems in these processes to establish identity trust~\cite{beheshti2020towards}. \item Retail: Face Recognition is used in the retail industry to identify shoplifters or premium customers as they enter the store and quickly notify the manager~\cite{beheshti2020towards,elahi2021recommender}. 
                \item Education: In education, face Recognition is used in automated online proctoring, class attendance checking, school facilities access control and security reasons~\cite{shabani2022icreate,wang2021assessment2vec}.
                \item Public Transport: In public transport, face recognition could be used to track the capacity and population, identify the wanted person by police request and alarm if there is a medical emergency~\cite{ashraf2020novel,farhood2022recent}. 
            \end{itemize}
    
        \subsubsection{Industry-Based Presentation Attack Detection Approaches}
        \label{section:industry}
            This part introduces the leading companies that have built presentation attack detection approaches. We analyse the input and test types in terms of being intrusive (active) or non-intrusive (passive). There is a lack of information regarding the details of the methodology they are using, but almost all of them claim to use various neural networks and deep learning algorithms. Table~\ref{table:company} summarises our investigation of the companies providing a PAD approach. 
        
            \begin{longtable}[c]{lll}
                \caption{Liveness detection approaches in the industry}
                \label{table:company}\\
                \toprule
                \textbf{Company Name} & \textbf{Active/ Passive}                                  & \textbf{Input Type} \\
                 \midrule
                \endhead
                Aware
                ~\cite{aware_2022}                     
                & Active/Passive                       & -                   \\
                AWS
                ~\cite{liveness_detection_framework_1991}
                & Active                               & Selfi Video         \\
                BioID
                ~\cite{administrator_2022}             
                & Active                               & Selfi Video         \\
                ElectronicID
                ~\cite{elecID}                  
                & Active                               & Selfi Video         \\
                Facetec
                ~\cite{facetec}                      
                & Active                               & Selfi Video         \\
                IDcentral
                ~\cite{idcentral}                  
                & Passive                              & Single Image       \\
                Idrnd
                ~\cite{Idrnd}                          
                & Passive                              & Single Image        \\
                Luxand
                ~\cite{Luxand}                        
                & Passive                              & Single Image        \\
                SpoofSence
                ~\cite{SpoofSence}                
                & Passive                              & Single Image        \\
                Sumsub
                ~\cite{Sumsub}                        
                & Active                               & Selfi Video         \\
                \bottomrule
            \end{longtable}
    
    \subsection{Datasets}
        This section describes the databases that were used in the primary studies. Table~\ref{table:datasets} depicts 15 benchmark face spoofing datasets. We extract various details from their descriptions. The information includes the year the dataset was created, the total number of samples used in the dataset, the number of subjects (individual faces), the format of data (whether it is a video or a photo), the spoofing attacks contained in the dataset (we divide the attacks into three categories and add three columns of photo attack, video attack, and 3D attack in this regard). We also incorporate varied conditions, such as whether the samples are collected in different illumination settings.
        
        \begin{longtable}[c]{llllllll}
            \caption{Benchmark Datasets}
            \label{table:datasets}\\
            \toprule
            \textbf{Name} &
              \textbf{Year} &
              \textbf{Samples} &
              \textbf{Subjects} &
              \textbf{Format} &
              \textbf{\begin{tabular}[c]{@{}l@{}l@{}}Photo, Video,\\Mask Attacks\end{tabular}} &
              \textbf{\begin{tabular}[c]{@{}l@{}}Different \\ Illumin-\\ation\end{tabular}}  \\ 
              \midrule
            \endhead
            \begin{tabular}[c]{@{}l@{}}Yale Face\\~\cite{yalefacedatabase} \end{tabular}                            & 2001 & 5760  & 10   & \begin{tabular}[c]{@{}l@{}}60-Frame\\ Video\end{tabular}  & Y, N, N & N \\
            NUAA~\cite{tan2010face}                          & 2010 & 9123  & 15   & \begin{tabular}[c]{@{}l@{}}500-Frame\\ Video\end{tabular} & Y, N, N & Y \\
            \begin{tabular}[c]{@{}l@{}}Print Attack\\~\cite{anjos2011counter}  \end{tabular}                        & 2011 & 400   & 50   & \begin{tabular}[c]{@{}l@{}}9-Second\\ Video\end{tabular}  & Y, N, N & Y \\
            \begin{tabular}[c]{@{}l@{}}Replay Attack\\~\cite{chingovska2012effectiveness} \end{tabular}& 2012 & 1300  & 50   & \begin{tabular}[c]{@{}l@{}}9-Second\\ Video\end{tabular}  & Y, Y, N & Y\\
            \begin{tabular}[c]{@{}l@{}}CASIA FASD\\~\cite{zhang2012face} \end{tabular}                 & 2012 & 600   & 50   & Video                                                     & Y, Y, N & Y  \\
            3DMAD~\cite{nesli2013spoofing}                                 & 2013 & 255   & 17   & \begin{tabular}[c]{@{}l@{}}300-Frame\\ Video\end{tabular} & N, N, Y & Y \\
            \begin{tabular}[c]{@{}l@{}}MSU-MFSD\\~\cite{wen2015face} \end{tabular}                             & 2015 & 440   & 55   & Video                                                     & Y, Y, N & N  \\
            \begin{tabular}[c]{@{}l@{}}HKBU-MARs\\~\cite{liu20163dd} \end{tabular}                  & 2016 & 1008  & 14   & \begin{tabular}[c]{@{}l@{}}10-Second\\ Video\end{tabular} & N, N, Y & Y  \\
            \begin{tabular}[c]{@{}l@{}}MSU USSA\\~\cite{patel2016secure}  \end{tabular}                & 2016 & 9,000 & 1000 & Video                                                     & Y, Y, N & Y \\
            \begin{tabular}[c]{@{}l@{}}Replay Mobile\\~\cite{costa2016replay}\end{tabular}                          & 2016 & 1190  & 40   & \begin{tabular}[c]{@{}l@{}}10-Second\\ Video\end{tabular} & Y, Y, N & Y \\
            \begin{tabular}[c]{@{}l@{}}OULU-NPU\\~\cite{boulkenafet2017oulu}\end{tabular}                         & 2017 & 5940  & 55   & Video                                                     & Y, Y, N & Y \\
            SiW~\cite{liu2018learning}                                    & 2018 & 4478  & 165  & \begin{tabular}[c]{@{}l@{}}15-Second\\ Video\end{tabular} & Y, Y, N & Y \\
            \begin{tabular}[c]{@{}l@{}}ROSE-Youtu\\~\cite{li2018unsupervised}   \end{tabular}                         & 2018 & 4225  & 25   & \begin{tabular}[c]{@{}l@{}}10-Second\\ Video\end{tabular} & Y, Y, Y & Y \\
            WFFD~\cite{jia2019database}                      & 2019 & 4400  & 740  & Photo                                                     & N, N, Y & N \\
            \begin{tabular}[c]{@{}l@{}}CASIA SURF\\~\cite{zhang2020casia}\end{tabular}                       & 2020 & 21000 & 1000 & Video                                                     & Y, N, N & N \\
            \bottomrule
        \end{longtable}
    
    \subsection{Challenges}
    \label{section8:challanges}
        The variety of attack types, the use of RGB image/video of varying low to high resolutions, technological improvements and the invention of new high-resolution attacks, and the requirement of deploying a robust PAD system in high-security domains make presentation attack detection a highly challenging task. As a result of examining various studies on PAD, in this section, we can highlight some of the crucial challenges of the PAD process by reviewing previous academic studies and observing industry activities.  
        Dataset collection is one of the most common challenges in PAD. PAD methods require different data. Some require specific user interaction, while others require a specific image quality, a specific amount of frames, and other requirements. As a result, finding an available dataset might be difficult, and in some cases, creating one's dataset is the best option. On the other hand, creating a diverse and large dataset is time-consuming. Some attacks are difficult to replicate, such as a dataset including videos of super-realistic wax figures in various dimensions and with different camera resolutions.
        
        Another challenge for PAD approaches is that new forms of more realistic attacks will emerge as technology improves. Hence, one challenge that has attracted attention in recent years is developing a system that can cover both seen and unseen attacks. Since most PAD methods are resilient to specific attack types, the best approach to establishing a robust PAD system for all attacks is to ensemble different PAD methods. Another PAD challenge is building a fusion model, as there are numerous fusion methods, and selecting the optimum one to ensemble different PAD techniques is challenging.
        Another possible challenge in some PAD applications is when the approach needs to be run on a limited resource (CPU, GPU and Memory). Most highly accurate PAD models use complex and heavy-weighted deep-learning approaches that are resource intensive. Building a CPU and Memory-light yet a robust and accurate PAD approach is challenging.
        The last challenge is that all researchers conducted their experiments on relatively controlled and low-diversity datasets. The PAD system will deal with various camera resolutions, lighting conditions, backgrounds, and facial expressions and poses in real-world scenarios. Another significant challenge is that the studied methods might only sometimes work well with real-world cases.
    
    \subsection{Summary}
        We conducted a comprehensive literature review of the end-to-end pipeline of presentation attack detection approaches and various forms of presentation attacks in this study. We also addressed selecting the best method for a specific attack, evaluation metrics and criteria, and PAD challenges. We also comprehensively discussed the real-world industry applications of presentation attack detection. In addition, we offered a comprehensive list of available datasets in this field.


\section{Methodology}
\label{chap:3}
        
    This Section comprehensively explains our proposed active presentation attack detection method and framework. Section~\ref{chap:2} extensively discussed the academic and industrial approaches for presentation attack detection. Also, an introduction to multiple types of presentation attacks, including 2D and 3D attacks and an overview of the existing challenges of developing presentation attack detection methods were given. In this work, we focus on 2D presentation attack detection by introducing a novel efficient active PAD approach. Our approach addresses some of the mentioned challenges (it is CPU-light, requires no training data, and works with low-quality frames), and it is both user-friendly and robust to 2D attacks.
    We also suggested a framework for online exam proctoring in addition to this method. In the past two years, online education and exams have expanded significantly worldwide due to COVID-19, and many exams have been administered online at schools and universities. We propose a framework of end-to-end online exam proctoring followed by a mock-up admin-side application using this motivation scenario as one of the potential uses of presentation attack detection.
        
    \subsection{Introduction to the Proposed Active Test}
        
\begin{figure}[htbp]
            \centering
            \includegraphics[width=\linewidth]{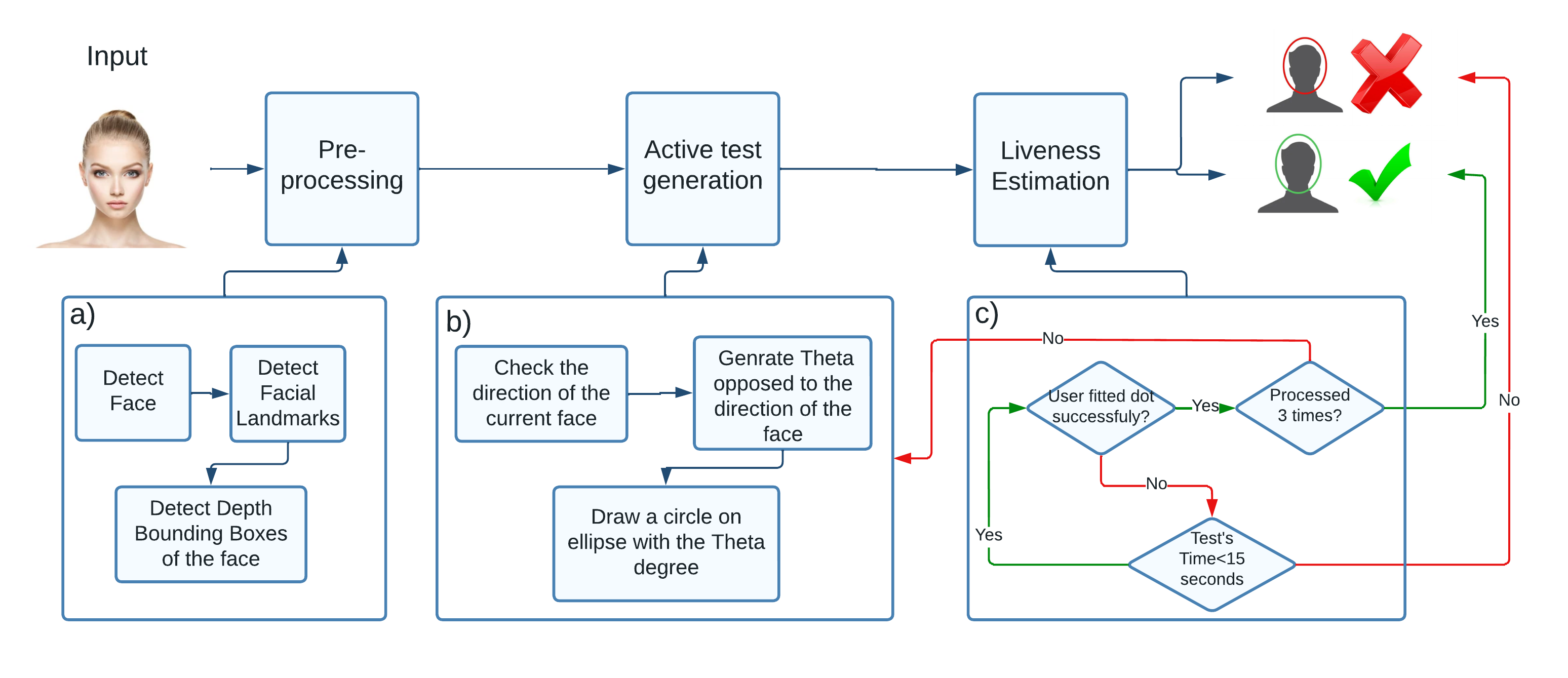}
            \caption{The active presentation attack detection service architecture.}
            \label{fig:architecture}
            
        \end{figure}

        We widely analysed different presentation attack detection methods in Section~\ref{chap:2}. These methods have been developed in academia and industry, utilising deep learning and machine learning techniques. According to Section~\ref{section8:challanges}, Some of the challenges for these models are as follows:
            \begin{itemize}
                \item Data Collection: Presentation attack types are diverse. Different 2D and 3D attacks exist, and also as the technology grows, new attacks are being generated, such as deepfake. Furthermore, reproducing some attacks requires access to unique technologies and materials, such as super-realistic wax figures. Hence, collecting a sufficient amount of all PA types is exceptionally challenging.
    	        \item Data Quality: The quality of camera sensors in different PAD applications varies widely. Most currently existing models require a specific minimum quality to detect PAs, such as those focusing on the texture or life signs like heart rate. This limitation restricts these models from being used in all of the applications.
    	        \item Resource Intensity: Most presentation attack detection techniques now use deep learning techniques and therefore require significant computational resources. This creates restrictions for some PAD uses with fewer computational resources.  
            \end{itemize}
        As the primary contribution of this paper, we introduce a novel, efficient active PAD service that addresses the challenges mentioned above. Our approach does not require training data, can process low-quality images, and is CPU-light. In addition to resolving these issues, we have tested this service with various users and found it both robust to 2D PAs and user-friendly. Figure~\ref{fig:architecture} depicts the service architecture. Pre-processing, Active Test Generation, and Liveness Evaluation are the three major components of the procedure.

    \subsection{Pre-processing}
        As discussed, in this work, we focus on detecting 2D PAs, which are most common since they do not require special equipment and are highly effortless to produce. Two of the features of a live face that could be utilised to detect 2D attacks are spatial features and motion, as previously discussed in Section~\ref{chap:2}. In this work, we use these features to detect 2D attacks. 
        During pre-processing, the face and its facial landmarks are first detected. In order to detect faces and landmarks, multiple approaches are available. Dlib (\url{http://dlib.net/face_landmark_detection.py.html}), ML Kit's face detection (\url{https://firebase.google.com/docs/ml-kit/detect-faces}), Viola-Jones~\cite{viola2001} and Mediapipe (\url{https://google.github.io/mediapipe/solutions/face_mesh.html}) are some popular packages and services. We use the Mediapipe Face Mesh approach in this work, which uses "lightweight model architectures together with GPU acceleration throughout the pipeline"~\cite{mediapipe}. This will give us a real-time and lightweight approach for the first step of our pre-processing. Mediapipe's output would be the detected faces' bounding boxes and 468 landmarks for each face. Although we only utilise 5 landmarks in our active PAD approach. 
        
\begin{figure}[htbp]
            \centering
            \subfloat[]{\includegraphics[width=.4\linewidth]{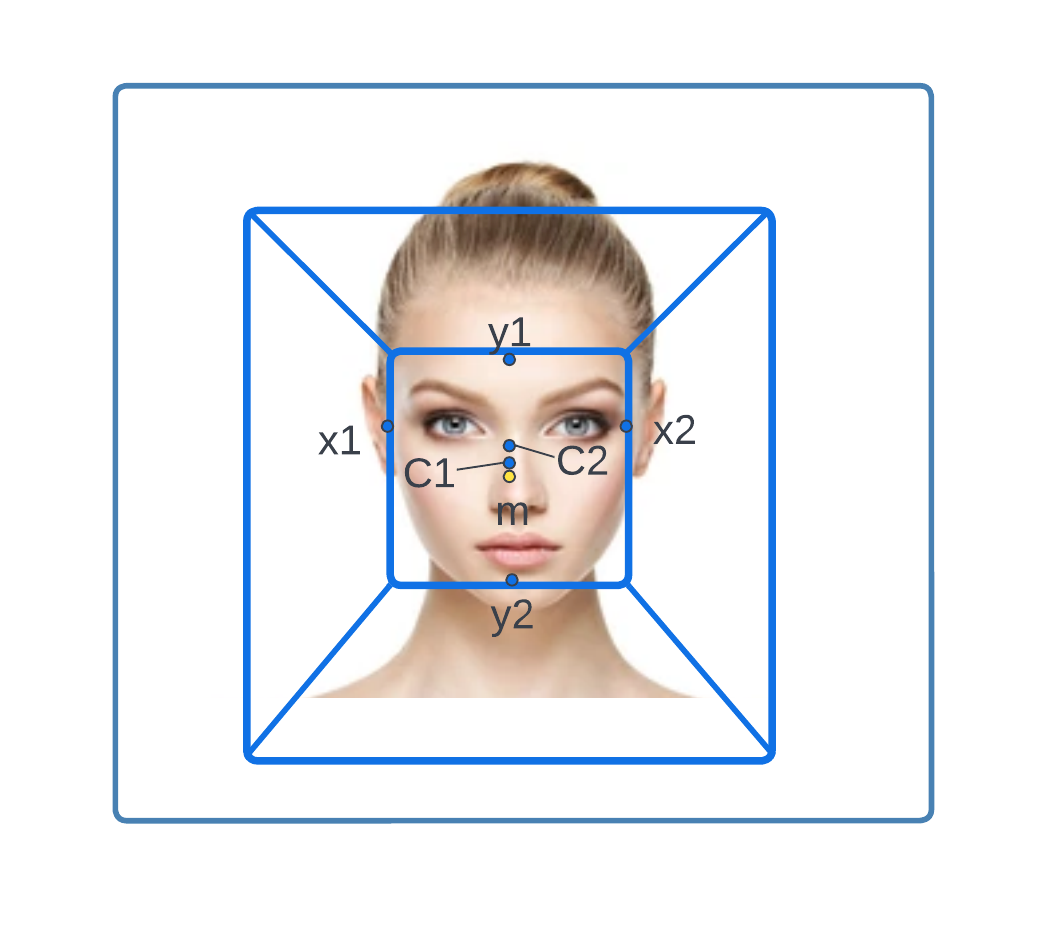}} 
            \subfloat[]{\includegraphics[width=.42\linewidth]{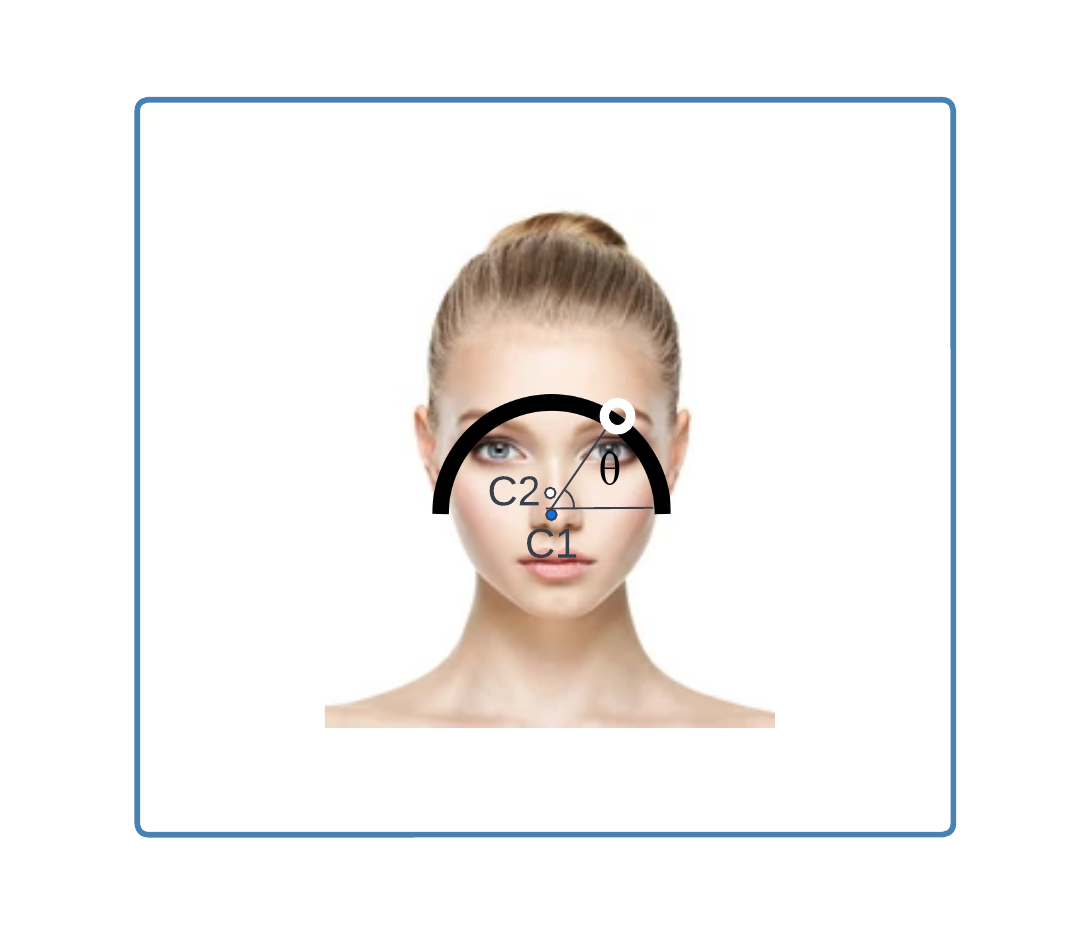}}
            \caption {The demonstration of the active test generation logic: a) The preprocessing of the active test, including face detection, facial landmark detection and depth bounding box creation; and b) The dot and circle creation logic.} 
            \label{fig: demonstration_liveness} 
        \end{figure}

\begin{figure}[htbp]
            \centering
            \includegraphics[width=0.5\linewidth]{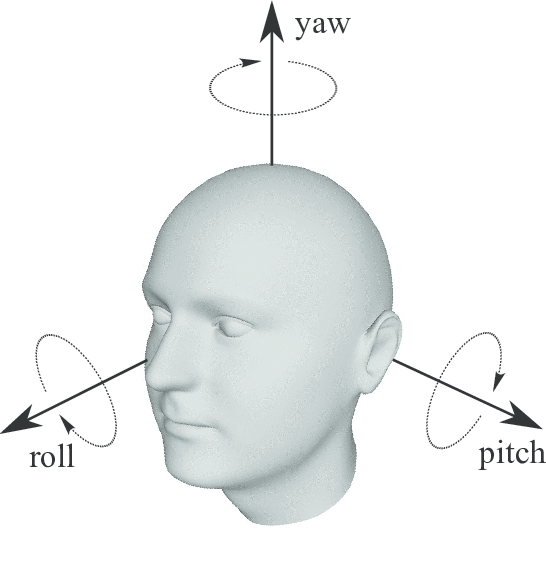}
            \caption{Different head poses~\cite{fernandez2016driver}.}
            \label{fig:head_pose}
        \end{figure}

        As illustrated in Figure~\ref{fig: demonstration_liveness}~(a), The face bounding box (the inner rectangle to the centre of \begin{math}c1\end{math}) and five landmarks of (i) \begin{math} x1\end{math}: x-axis of the left side of the face; (ii) \begin{math} x2\end{math}: x-axis of the right side of the face; (iii) \begin{math} y1\end{math}: the y-axis of the middle of forehead; (iv) \begin{math} y2\end{math}: the y-axis of the middle of the chin; and (v) \begin{math} m\end{math}: the x and y-axis of the tip of the nose are extracted from Mediapipe API. 
        In order to get the spatial features of the face, we create the outer box (Pose-determining Box) of Figure~\ref{fig: demonstration_liveness}~(a). This box reflects the pose of the head determined by two angles: yaw (horizontal movement) and pitch (vertical movement), Figure~\ref{fig:head_pose}. 
        The pose-determining box to the centre of \begin{math}c2\end{math} is created based on a mathematical operation of the observed landmarks:  if \begin{math} d_l\end{math} is the length between the outer left side and the tip of the nose and \begin{math} d_r\end{math} is the length between the outer right side and the tip of the nose, \begin{math}c2 \end{math} is defined as the \begin{math}c1\end{math} plus \begin{math} d_l - d_r \end{math}. The same logic also applies to vertical motion. Algorithm~\ref{alg:outerbox} shows the calculation of the centre of the pose-determining box. Just like the pose-determining box, its centre (\begin{math}c2\end{math}) will also reflect the pose of the head, and therefore the head pose could be outlined by tracing \begin{math}c2\end{math}.

        \begin{algorithm}
            \caption{Pose-determining box generation.}
            \label{alg:outerbox}
            \begin{algorithmic}[1]
                \State $bbox,landmarks \gets mediapipe(frame)$
                \State $c1 \gets Center Of Bbox$
                \State $d_l \gets abs(landmarks.x1-landmarks.m.x)$
                \State $d_r \gets abs(landmarks.x2-landmarks.m.x)$
                \State $d_u \gets abs(landmarks.y1-landmarks.m.y)$
                \State $d_d \gets abs(landmarks.y2-landmarks.m.y)$
                \State $horizontalPose \gets d_l - d_r$
                \State $verticalPose \gets d_u - d_d$
                \State $c2 \gets c1 + (horizontalPose,verticalPose)$
            \end{algorithmic}
        \end{algorithm}
             
    \subsection{Active Test Generation}
            Active PAD is a user-interactive method. Numerous active PAD approaches require the user to do simple tasks such as turning their head, smiling and blinking, speaking randomly generated numbers, and making random faces~\cite{liveness_website}. These tests are designed to be user-friendly but are susceptible to specific 2D attacks. E.g., smiling or turning head to the left or right are some common actions on most videos, and these videos could be used to spoof the system. Hence, these tests are mainly used as a component of a more extensive system that performs more complex deep-learning approaches. In this work, we offer a novel active PAD service that relies purely on the user's interaction, with no further deep learning/ machine learning processes. It solves the problem with the aforementioned tests by tracking the head and asking users to move their head to an exact position in a straightforward and user-friendly manner. The exact position is randomly generated for each test. Hence, it is a low chance of the system being spoofed by a video with the head moving exactly towards those randomly generated points. This section provides a detailed discussion of our proposed approach.
             
            We introduce a novel approach to navigate the user to pose their head towards a specific point. With the purpose of doing so, we present a circle and a dot on the screen and ask the user to fit the dot into the circle by moving their head. The dot's centre is \begin{math}c2\end{math}, which moves based on and in the same direction as the head movements. The circle is also created on a half ellipse on the face's bounding box. As shown in Figure~\ref{fig: demonstration_liveness}(b) and indicated in Algorithm~\ref{alg:active_test}, a half ellipse to the centre of \begin{math}c1 \end{math}, and the width and height of \begin{math}a \end{math} and \begin{math}b \end{math}, respectively, is first assumed in the upper side of the face bounding box. Then a $\theta$ value is randomly defined between \begin{math}0 \end{math} and \begin{math}\pi \end{math}, describing the circle's position on the half ellipse. To improve the service's robustness, the $\theta$ value is defined in the opposite direction of the current head position, as indicated in Algorithm~\ref{alg:active_test}. The user is then given the dot to the centre of \begin{math}c2\end{math} and asked to move their head to position the dot within the circle created on the ellipse.
             
            \begin{algorithm}[htbp]
                \caption{Theta, dot and circle creation.}
                \label{alg:active_test}
                \begin{algorithmic}[1]
                    \If{$c1.x\geq c2.x$} 
                        \State $theta \gets random(((3 \times pi) / 2), 2 \times pi)$
                    \Else
                        \State $theta \gets random(pi, ((3 \times pi) / 2))$
                    \EndIf             
                    \State $timer \gets startTimer()$ 
                    \While{$timer \leq 15$ }
                        \State $frame \gets getNewFrame()$ 
                        \State $a \gets \frac{abs(landmarks.x1-landmarks.x2)}{2}$ 
                        \State $b \gets \frac{abs(landmarks.y1-landmarks.y2)}{2}$ 
                        \Comment{Half Ellipse's width of $2\times a$ and height of $2\times b$:}
                        \State $circle \gets (a\times cosine(theta) + c1.x, b\times sine(theta) + c1.y)$
                    \EndWhile
                \end{algorithmic}
            \end{algorithm}

    \subsection{Liveness Evaluation}
        Multiple tries and errors revealed that the average live user could complete the task in less than five seconds. To make it more robust to 2D video attacks that may involve head movement in specific directions, the circle's position would be updated, so the user must fit the dot to the circle three times in a row. Therefore, there we utilise a 15-second timer for the task. The user passes the PAD test if they correctly align the dot with the circle three times within the time limit; otherwise, they fail. To check if the user has fit the dot to the circle, we assume a radius \begin{math}r\end{math} for the circle and mathematically, we check that the coordinates of the point are inside the coordinates of the circle as indicated in Algorithm~\ref{alg:evaluation}. 
           
        \begin{algorithm}[htbp]
            \caption{Liveness evaluation.}
            \label{alg:evaluation}
            \begin{algorithmic}[1]
                \If{$(c2.x-circle.x)^2 + (c2.y-circle.y)^2 \leq r^2 $} 
                    \State $test \gets passed$ 
                \EndIf
            \end{algorithmic}
        \end{algorithm}
    
    \subsection{Online Exam Proctoring Framework}
        As explained in Section~\ref{section:industry}, PAD has several industrial applications and use cases. One of the mentioned applications of PAD is for online exam proctoring. The history of online exams dates back to early 2000. Online exams could be held in two ways of machine-based and human-based. In the first 10 years, all the online exams were being held human-based, but in late 2010, the early approaches of automatic and machine-based approaches came out. In the human-based form, the test candidates are linked to an exam supervisor via a computer and their video, and the exam supervisor invigilates them live. In the automated approaches, the machine captures the candidate's video and, using some AI-based technologies such as face recognition, facial expression detection, liveness detection, voice detection, and gaze tracking, supervises the exam and finds and logs the suspicious behaviours live~\cite{selwyn2021necessary}. 
        
        In the last two years, with the advent of COVID-19, more exams have been held online, and online exam proctoring tools have been widely used\cite{sankey2021covid}. There are several academic and industrial approaches for automated online exam proctoring\cite{hussein2020evaluation, atoum2017automated,kasinathan2022proctorex}. Most automated online exam proctoring methods are mainly designed for on-screen exams. We propose a new framework for automated online exam proctoring in this part. The exam could be held on-screen or on paper in this framework. We utilise the human-based online exam proctoring approach and offer a framework that could convert the human-based approach to its automated version. 
        
        \subsubsection{Framework Description}
        
            As previously mentioned, in a human-based approach, an exam supervisor is connected to the candidate and can observe their live video. At first, they ask the candidate to present their ID card and ensure the face match between the candidate and their ID card's photo. Then they ask the candidate to scan the room and the desk. Then they start the exam (on-screen or on paper). The proctor will fill out the report per suspicious action during the exam. We obligate this approach and propose a new automated online exam proctoring framework. As the candidates need to be checked in real-time, presentation attack detection plays a significant role in avoiding cheating using a presentation attack, such as presenting a replay video instead of a real user.
            
\begin{figure}[htbp]
                  \centering
                  \includegraphics[width=\linewidth]{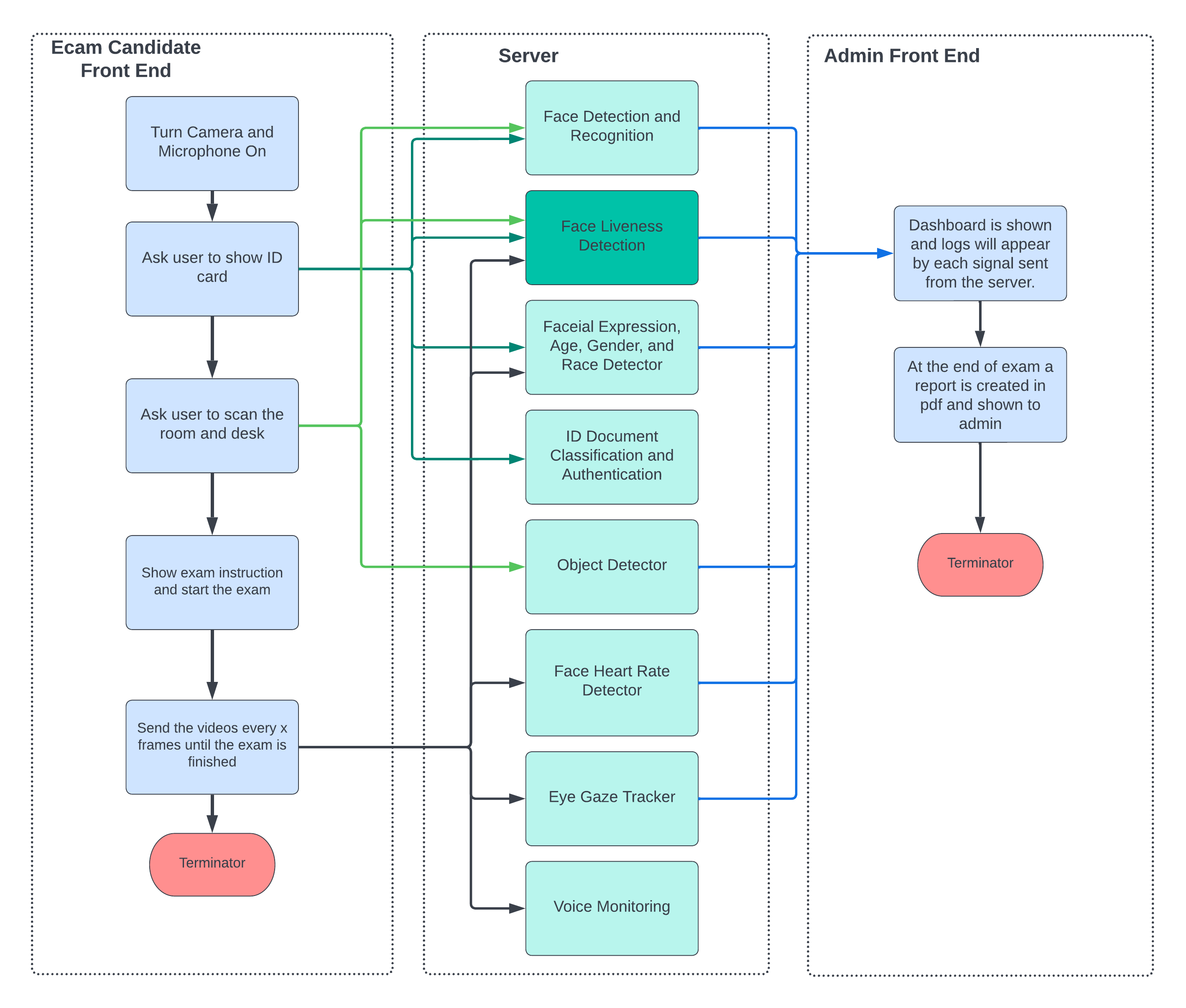}
                  \caption{The Proposed Online Proctoring Framework.}
                  \label{fig:online_exam}
            \end{figure}

            Figure~\ref{fig:online_exam} illustrates our proposed framework. When the exam candidate is connected, they need to turn on the video and microphone. Then instructions will be shown to the candidate, and ask the candidate to present their ID card to the camera. In this part, based on face recognition technology, a face match check will be done on the ID card's photo and the candidate's face. In this part, the identity document classification and authentication approach can extract and validate the information with the ones in the database. Then the user is shown instructions and asked to scan the room and desk. For this step, a configuration setting could be developed on the admin side, and based on the exam requirements, the admin could set the list of non-permitted objects. The system in this step uses this list and, with the object detection approaches, detects the forbidden objects, and if there are any, the system will ask the user to put that object away and re-scan the room and desk. The system will also use face-detection techniques to ensure no one else is in the room. Finally, the system shows the exam instructions, and the exam timer will start. From this point, facial expressions, liveness, heart rate, and eye gaze will be checked and tracked for all the frames. Any abnormal behaviour, such as a non-normal facial expression or unusual or not permitted sound, will be logged and sent to the admin. Finally, when the exam is finished, all the logs will be entered in a PDF template and sent to the admin. 
            
        \subsubsection{Discussion}
            The proposed framework is divided into multiple components, making the system development feasible using the available technologies and APIs related to each component. As already mentioned, most automated online exam proctoring approaches cover face detection and recognition, facial expressions detection, presentation attack detection, voice monitoring and eye gaze tracking. The novelty of our proposed framework is to add another component using one of the life sign-based presentation attack detection methods, heart rate detection. We offered this component not only regarding detecting the presentation attacks but as a separate new component to monitor the stress level and health of the candidate during the online exam. To the best of our knowledge, our framework is offering the use of heart rate and student health checks in an automated online exam proctoring for the first time. The significance of adding such a component to the system is that exams could be stressful for some candidates, and relying on an automated online proctoring solution could jeopardise student health. This is because the system is not able to offer the right solution in an emergency and can only log the status. The inability of the system to respond to these emergencies is one reason we designed our framework to log and notify the admin side of any abnormal behaviour in real-time. So, the rPPG signals could be used to detect the heart rate, and if the heart rate of an exam candidate was too high or low, the admin (a human proctor) could immediately connect to that candidate's session and check on their health state.


\section{Evaluation and Experiment}
\label{chap:4} 
    
    \subsection{Analyses of the Proposed Approach's Claims}
        We developed an efficient active presentation attack detection approach that requires no training data, is CPU-light, and works with low-quality camera resolutions. At the same time, it is highly robust to 2D presentation attacks and user-friendly. The performance of the system and its degree of usability are both covered in the evaluation section. In this section, we discuss the first three claims by accomplishing experiments and mathematical proofing. 
           
        \subsubsection{Baselines}
        To better understand how our system covered the gaps in the literature, some baselines are introduced. We selected the baselines from three categories of methods used to detect 2D attacks: Deep learning approaches supervised to extract depth, motion-based approaches to detect Spatial features, and machine-learning approaches to detect texture. The similarities between deep learning and motion-based approaches and our approach are that all rely on the Spatial features of the face and are designed to detect 2D attacks. We also considered comparing two popular texture machine-learning approaches with our approach in terms of their data quality, collection and resource usage. The following are the selected baseline models:
        \begin{itemize}
            \item Texture-based machine learning algorithms:
                \begin{itemize}
                    \item LBP~\cite{maatta2011face}: A Local Binary Pattern is a pixel-level texture feature descriptor that extracts feature vectors from the images. Section~\ref{section:texture} describes the Local Binary Patterns and the presentation attack detection using this methodology in detail. 
                    \item IQM~\cite{galbally2013image}: Image Quality Metrics are some of the mathematical calculations done on the image pixels to assess the image quality, such as Mean Square Error, Gradient Magnitude Error, and Laplacian Mean Square Error. To detect the presentation attacks, since it is believed that the quality of the presentation attacks is less than the genuine face, multiple IQM calculations are used to create a feature vector. The feature vector is then trained with a classifier, such as SVM. 
                \end{itemize}
            \item Motion-based machine learning algorithms:
                \begin{itemize}
                    \item Optical Flow~\cite{kollreider2009non}: Optical Flow is the process of tracing a point in multiple frames of a video. In motion analysis, this method is used to trace the face's motion and background. Kollreider et al.~\cite{kollreider2009non} use two regions of the face (ears and nose) and ask the user to move their head to the sides. Then, it detects the 2-dimensional presentation attacks by tracing the movements of these two areas. More detail is given in Section~\ref{section:motion}.
                    \item Invariant Cross Ratio~\cite{de2012moving}: This work uses a mathematical lemma to detect 2D presentation attacks. The lemma states that if we have five points in a flat surface and no three of them are in a line, then some motion-invariant cross ratios exist, and the ratio will not change if the surface moves. Marsico et al.~\cite{de2012moving} use this lemma and select five points on the face (e.g. nose, eyes corners, and ears corners) and then ask the user to move their face. If the cross ratios change, the points are not on the same plane surface, and hence the shape of the face is 3D. More detail is given in Section~\ref{section:spatial}
            \end{itemize}
            \item Deep learning depth map-based algorithm:
                \begin{itemize} 
                    \item Auxiliary~\cite{liu2018learning}: This approach uses a Convolutional Neural Network architecture followed by a Recurrent Neural Network architecture. The architecture is designed to create the face's depth map from the input frames. If the depth map shows no depth around the face area, it detects that frame as fake, and if the depth map of the face is entirely created, it detects the frame as live.
                \end{itemize}
        \end{itemize}
            
        \subsubsection{Data Collection and Quality}
            We have tested our approach and the minimum required quality for our approach to work properly is \begin{math}60\times60\end{math} pixels 4Kb images (with a face resolution of \begin{math}20\times36\end{math} pixels). Comparing with the baselines, LBP and IQM use normalized face images of \begin{math}64\times64\end{math} pixels,  Optical Flow uses \begin{math}300\times240\end{math} pixels, Invariant Cross Ratio uses \begin{math}640\times480\end{math}, and Auxiliary uses \begin{math}256\times256\end{math}. In this comparison, the minimum resolution required in our approach is significantly less than that of deep learning and motion-based approaches. However, the texture-based approach works as well as our approach on low-quality frames.
            
            Also, our approach does not use machine/deep learning in the backend process, so it does not require any training data. As long as the presented face artifact cannot move its head to multiple positions asked by the application, the test will perform resilient against it. However, the rest of the baseline approaches except for the Invariant Cross Ratio need a training stage. Also, a threshold needs to be defined in the Invariant Cross Ratio approach and finding the best threshold also needs a training stage. This is a significant advantage of our proposed approach compared to the baselines. 
        
        \subsubsection{Resource Intensity}
           The system architecture and the Algorithms in Section~\ref{chap:3} show that to perform our active test, we are using some simple mathematical operations, which makes the service CPU-light. To prove it, we compare the number of operations with different baselines. Assuming LBP as a texture-based baseline, disregarding the preprocessing, Local Binary Pattern (LBP) has been applied on \begin{math}64\times64\end{math} pixels image and then its histogram is fed to an SVM. To apply LBP, 8 multiplication and seven summation operations are done for each pixel. In total \begin{math}62\times62\times15=57660\end{math} operations are done. And then, for the SVM, at least 255 multiplications are done. In this work, less than 50 operations are done for each frame. Assuming the whole test takes 15 seconds and 20 frames are processed in each second, the number of operations will be \begin{math}20\times 15\times50=15000\end{math}, which is the worst-case scenario. Deep learning models use CNN architectures and have at least three layers with multiple kernels, making a higher time complexity than the baseline and our service. Regarding resource usage, the motion-based models are as CPU-light as our proposed approach. However, they still have an extra training phase compared with our proposed approach.
    
    \subsection{Evaluation}
        In this section, we will evaluate our work in terms of its robustness to the 2D attacks and usability. For the first part, since our approach depends on the user's response, no publicly available dataset could be used for testing, and we invited seven participants to test the system. For the second part, we created a trial desktop application of the proposed approach and provided a five minutes video describing the methodology and demonstration of the application. Then we created a Questionnaire and asked 23 participants to watch the video and understand the intent of our approach, then install and try the application on their PCs and finally fill out the Questionnaire. The experimental setups and the results of both parts are given in the following sub-sections.

        \subsubsection{Robustness}
            In this section, we test the proposed active presentation attack detection approach against various 2-dimensional attacks and analyse the results of our tests.
            \par
\begin{figure}[htbp]
                    \centering
                    \includegraphics[width=0.7\linewidth]{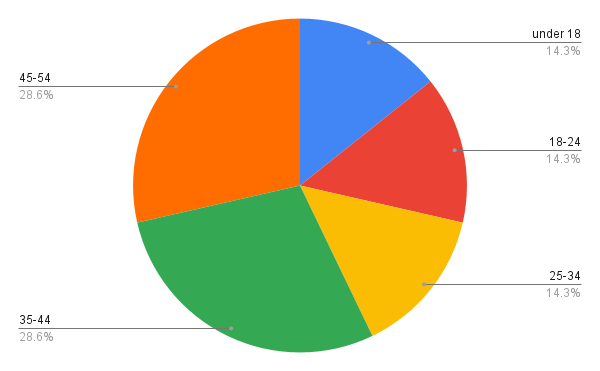}
                    \caption{The age range of the participants for testing the robustness of the system.}
                    \label{fig:age range robustness}
                \end{figure}

{Experimental Data}
                Most presentation attack detection studies use one or multiple public datasets to train/test their models. However, since our approach needs a user's live interaction, pre-collected data could not be used to test the systems. Hence, we invited seven participants to test the system with genuine and 2-dimensional fake attempts. The motivation scenario of our test is access control, and the audience of access control is the general public of all ages. Hence, the participant selection was not limited to a specific group, and only age distribution was considered. We tried to have a young age range, our most common audience. Figure~\ref{fig:age range robustness} depicts the age range of the participants. 
    
                To ensure the quality and quantity of our testing, we follow iBeta~\cite{ibeta_2022}, a software testing company that tests the PAD approaches and certifies their software. It uses ISO 30107-3\cite{standard2017information,ibeta_2022}, the presentation attack detection test standard. To test 2-dimensional attacks and paper masks attack, they test the system with 150 fake and 50 genuine samples using six subjects. To follow similar instructions, each of the seven participants tested our approach with 15 genuine and 18 fake attempts. Overall, 231 tests were done, 105 live tests and 126 fake tests.
                
            \par{Experimental Setup}
                The proposed approach is developed in python 3.8, using OpenCV-python 4.6.0 for grabbing the frames and Mediapipe 0.8.11 for preprocessing. A windows based trial application is also deployed to share with the participants. In this test, we used 4 computers (Macbook Pro M2, ASUS Zenbook UX303L, DELL Inspiron 5593, and ASUS Vivobook R424F) and seven participants. To cover all types of 2D attacks, each participant tested six printed photos (hard or soft), six digital photos (with different resolutions), and six digital videos (including a face with proper movements to spoof the system). Also, each participant tested the system with a live face 15 times. Table~\ref{tab:number of samples} shows the exact number of fake and live samples.
                
                \begin{table}[htbp]
                    \caption{The number of samples used for testing the system's accuracy}
                    \label{tab:number of samples}
                    \begin{tabular}{cccccc}
                        \hline
                        \multicolumn{6}{c}{\textbf{Number of Samples}}                           \\ \hline
                        Total & Live & Total Fake & Printed Photo & Digital Photo & Replay Video \\
                        231   & 105  & 126        & 42            & 42            & 42           \\ \hline
                    \end{tabular}
                \end{table}
                
            \par{Experimental Results}
                As extensively discussed in Section~\ref{section: evaluation metrics}, In 2016, an ISO Standard~\cite{iso_2021} for the Biometric Presentation Attack Detection evaluation metrics was defined. The terms Attack Presentation Classification Error Rate (APCER), Bona-fide Presentation Classification Error Rate (BPCER) and Average Classification Error Rate (ACER) were defined and corresponded to FAR, FRR and HTER, respectively. In this study, we will use these three metrics as well as accuracy to evaluate our system.
                
                Table~\ref{tab:eval results} shows the Accuracy, APCER, BPCER, and ACER of the proposed approach. Among 231 samples, the system reaches 99\% accuracy with zero BPCER and 1.5\% APCER. Our analysis of the false positive ones shows that all the false positive errors were regarding the videos with enough and proper motion. The system reaches 100\% accuracy among all types of photo attacks. This shows that the system can be improved in detecting video attacks by changing the number of times that the user needs to fit the dot into the circle or by changing the timeout period. Although the test results cannot be compared with the baselines reasonably since they cannot be tested on the same dataset, the results of some of the baselines that used the same evaluation metrics on their datasets are also given in Table~\ref{tab:eval results}. 
                
                \begin{table}[htbp]
                    \caption{The APCER, BPCER, ACER, and Accuracy of our approach and 3 baselines}
                    \label{tab:eval results}
                    \begin{tabular}{cccccc}
                        \hline
                        \textbf{Study}                                                   & \textbf{Dataset}                                              & \textbf{APCER\%} & \textbf{BPCER\%} & \textbf{ACER\%} & \textbf{Accuracy\%} \\ \hline
                        \begin{tabular}[c]{@{}c@{}}Our proposed \\ Approach\end{tabular} & Self-created                                                  & 1.58             & 0                & 0.79            & 99.13               \\
                        Auxiliary                                                         & \begin{tabular}[c]{@{}c@{}}OULU-NPU\\ Protocol 4\end{tabular} & 9.3 ± 5.6        & 10.4 ± 6.0       & 9.5 ± 6.0       & -                   \\
                        IQM                                                              & Replay Attack                                                 & 12.5             & 17.9             & 15.2            & -                   \\ \hline
                    \end{tabular}
                \end{table}       
        
        \subsubsection{Usability}
            To evaluate the system's usability and how easy it is for people in different age ranges, we conduct a user study, aiming to validate our hypotheses regarding the approach being user-friendly for a public audience. We have three primary hypotheses:
            \begin{itemize}
                \item H1. The system is user-friendly for all age ranges and public audiences.
                \item H2. Performing the test for live people is fast, and most people can perform the test between five to 10 seconds. 
                \item H3. The system could be used in the access control scenarios such as banking and phone unlock. 
            \end{itemize}
    
            \par
\begin{figure}[htbp]
                    \centering
                    \includegraphics[width=0.8\linewidth]{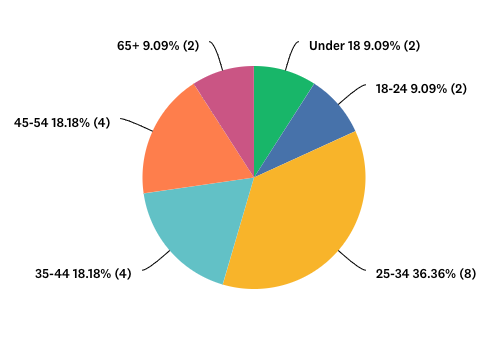}
                    \caption{The age range of the participants for testing the usability of the system.}
                    \label{fig:age range usability}
                    
                \end{figure}

{Experiment Setup}
                In this user study, we invited 23 participants to test our system. Similar to the robustness evaluation section, we use access control as the motivation scenario of our approach. Therefore, the participant selection was not limited to a specific group, and only age distribution was considered. However, since our Hypothesis 1 aims to evaluate the application's usability among all ages, we added more age ranges in the participant selection. Figure~\ref{fig:age range usability} demonstrates the age distribution of our 23 participants. 
                
                To train the participants, we created a five minutes video and windows based trial application of our approach\footnotemark{}\footnotetext{Link to the dropbox containing video and application: \url{https://www.dropbox.com/scl/fo/rh7grblr6ajbebiy0j5g9/h?dl=0&rlkey=f054da1okmnue6yc7tzqzb346}}. We asked them to watch the video to familiarise themselves with our approach and then install and run the application. For our user study to be fair, no additional comments and training were given for the test. Because if the test is going to be used as a liveness test in bank account access control, it should be easy enough to be learned on the first try. It cannot be used in such scenarios if it is not easy enough. Also, This time, we asked the users to assume that the service is highly accurate and robust to the 2-dimensional presentation attacks, test it with their live faces, and only focus on their user experience with the application. 
            
            \par{Questionnaire}
                We designed a Questionnaire\footnotemark{} \footnotetext{Link to the Questionnaire: \url{https://www.surveymonkey.com/r/P65X8CN}}with SurveyMonkey\footnotemark{} \footnotetext{Link to the website: \url{https://surveymonkey.com}} and asked the participants to fill it out based on their user experience with the application. In the survey, we designed six Questions to gather the required information to support our hypotheses. The first and second Questions, shown in Table~\ref{Questions}, are regarding H1. The first Question focuses on how easy it is to understand the app and how good the instructions are (the presentation of the dot and circle), and the second Question focuses on how easy performing the test is (fitting the dot to the circle for three consecutive times). The third Question discusses H2 by asking the user how fast the test is, and the fourth Question discusses H3 by asking the user's opinion on whether they prefer to use this application in the access control scenarios such as accessing a bank account. The fifth Question is asked to keep track of the age range and ensure that we have participants of all ages, and the last Question is for getting new ideas and feedback. Since the last Question was not mandatory, we got few yet beneficial feedback.
            
                \begin{center} 
                    \begin{table} 
                        \begin{tabular}{c c}
                            \includegraphics[scale = 0.4]{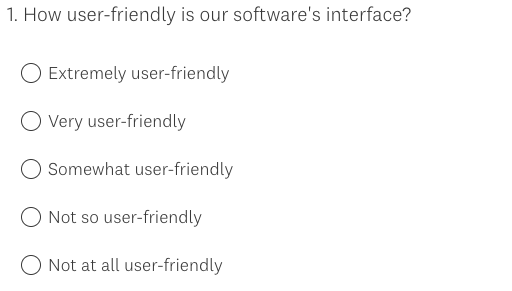} & \includegraphics[scale = 0.4]{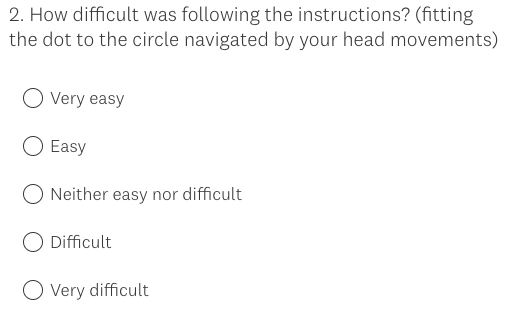} \\
                            (a) & (b) \\
                            \includegraphics[scale = 0.4]{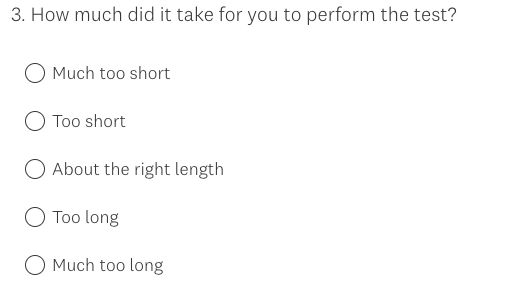} & \includegraphics[scale = 0.4]{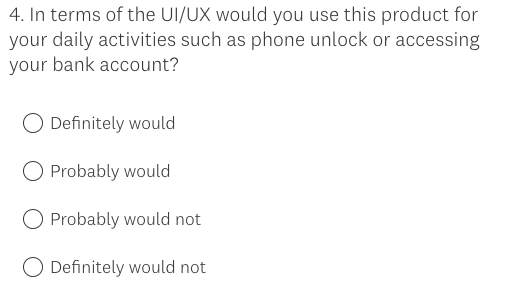} \\
                            (c) & (d) \\
                            \includegraphics[scale = 0.4]{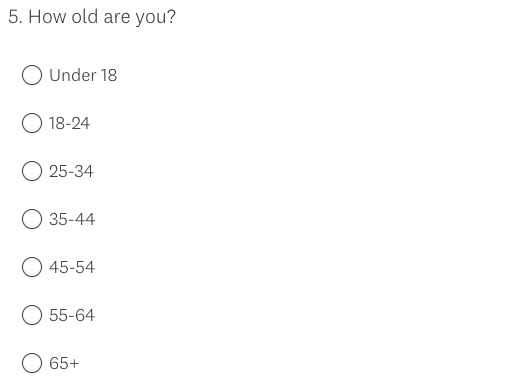} & \includegraphics[scale = 0.4]{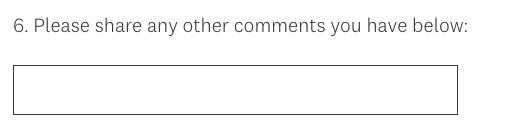} \\
                            (e) & (f) \\
                        \end{tabular} 
                        \caption{Questions of the Questionnaire: (a) The first Question is about the user interface and how easy to understand the test is;(b) The second Question is regarding how easy-to-use the application is; (c) The third Question is regarding how fast the participants could perform the test; (d) The fourth Question is regarding the application's usability in access control scenarios; and (e) the fifth Question is regarding the age distribution of the participants. (f) We also added a text box to gather general feedback and ideas.}  
                        \label{Questions} 
                    \end{table} 
                \end{center}
            
            \par
\begin{figure}[t]
                    \subfloat[]{\includegraphics[width=.5\linewidth]{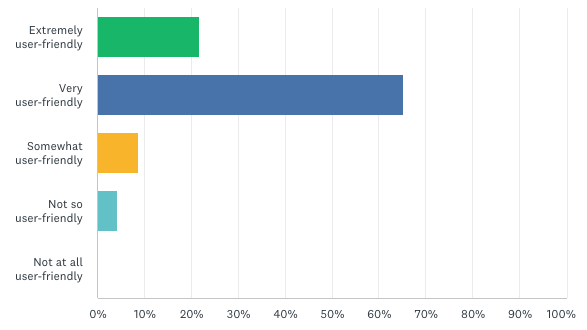}} 
                    \subfloat[]{\includegraphics[width=.5\linewidth]{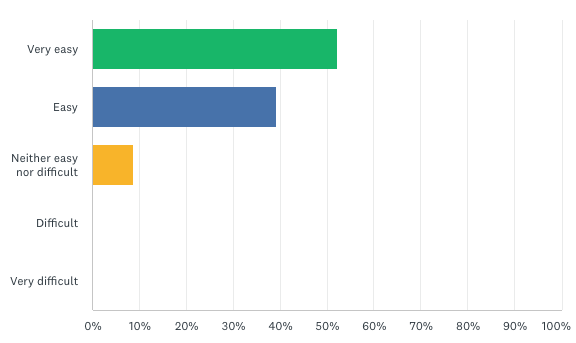}}
                    \caption {Results of the first and second Questions of the Questionnaire: a) Results of how user-friendly did participants find the test with the range of Extremely, Very, Somewhat, Not so, and Not at all user friendly; and b) Results of how easy-performing the test was with the range of Very Easy, Easy, Neither easy nor difficult, Difficult, and Very difficult.} 
                    \label{fig: results1} 
                \end{figure}

\begin{figure}[t]
                    \subfloat[]{\includegraphics[width=.5\linewidth]{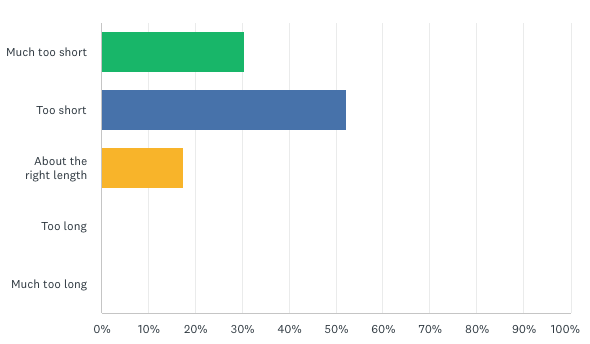}} 
                    \subfloat[]{\includegraphics[width=.5\linewidth]{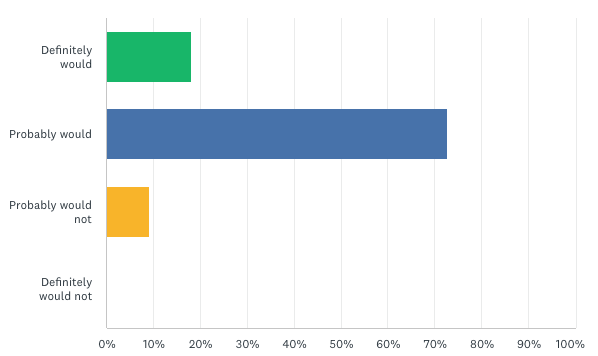}}
                    \caption {Results of the third and fourth Questions of the Questionnaire: a) Results of how fast participants find the test with the range of Much too short, Too short, About the right length, Too long, and Much too long; and b) Results of how functional the test is in the access control scenario with the range of Definitely would use the app, Probably would, Probably would not, Definitely would not .} 
                    \label{fig: results2} 
                \end{figure}

{Experiment Results and Discussion}
                We collected the Questionnaire answers for all 23 participants. The results of the first four Questions regarding our three hypotheses are demonstrated in Figures~\ref{fig: results1} and~\ref{fig: results2}. The following are the analyses of the hypotheses based on the results of our experiment:
                
                \begin{itemize}
                    \item Analyses of H1: As previously mentioned, the first two Questions demonstrate the participants' viewpoint on how user-friendly and easy-performing the test is. Figure~\ref{fig: results1}~(a) indicates that more than 80\% of the participants found the approach very user-friendly. Figure~\ref{fig: results1}~(b) depicts that almost 90\% of the participants found the instructions easy to follow. In general, the two first Questions show that almost more than 80\% of the participants agree with our claimed H1 Hypothesis, and our approach is user-friendly for all ages.
                    \item Discussion on the results of H1: Since we have a slight statistical society and from some of the age ranges, we only have one or two participants, we need to check the exact responses for not being user-friendly or easy-performing to ensure our conclusion is valid. The participants voting "Not so user-friendly or Somewhat user-friendly" age range is mostly 45-54. In this user study, we had four participants in this age range. The results show that almost half of the participants in this age range did not find the proposed approach relatively easy. However, the number of participants must be more significant to conclude with certainty. Nevertheless, we can conclude that the approach is undoubtedly user-friendly to young and early middle-aged adults. 
                    \item Analyses of H2: Question 3 of the Questionnaire addresses this Hypothesis. Since we have a timeout of 15 seconds in our approach for fake attempts, we wanted to ensure that for the live attempts, a shorter time, such as five to 10 seconds is enough for all ages. As shown in Figure~\ref{fig: results2}~(a), more than 80\% of the participants could perform the test in a short time, and less than 20\% were able to perform the test successfully within 15 seconds but not in less than five to 10 seconds.
                    \item Discussion on the results of H2: These results indicate that the H2 Hypothesis is valid and that 15 seconds timeout is a fair amount to perform the test for all ages.
                    \item Analyses of H3: Question 4 of the Questionnaire addresses this Hypothesis. This Hypothesis is created to ensure that the access control scenario is a fair use case for our proposed approach. As illustrated in Figure~\ref{fig: results2}~(b), the results indicate that more than 90\% of the participants believed they probably would use the app in the access control scenarios such as bank accounts, and less than 10\% probably would not use the app. We can infer that most participants were happy with the app's performance but unsure to use the application in the access control scenario. 
                    \item Discussion on the results of H3: The comment section (Question 6) was to get feedback and ideas for future work. However, we received some comments about Question 4 and our access control motivating scenario. Followings are some of these comments: \textit{"I feel that following the dot in case of urgently opening an app could be a bit slower process."}, \textit{"As I am a apple Face id user I think the Face id is very quicker and reliable than this so I can't choose this identifier instead of Face id but if I use desktop computers it might be good for bank accounts login or any account, if they had my face information on their servers"}, and \textit{"I am a bit hesitated to use it for accessing my bank account"}. These comments show that accessing a bank account might not be the best scenario given the five to 10 seconds duration of the test and the availability of alternative approaches. For future steps, other motivating scenarios such as CAPTCHA could be studied. 
                \end{itemize}

\section{Conclusion and Future Work}
\label{chap:5}
    
    In this Section, we conclude this paper by highlighting our contributions and discussing some of the potential future works to expand the ideas introduced in this paper.
    
    \subsection{Conclusion}
         Biometrics are the unique features of individuals that are used to detect their identities. Biometric authentication identifies people's identities using biometrics. Various biometric authentication services exist, such as face recognition, fingerprint recognition and iris recognition. Face recognition is used in multiple industries such as health, education, banking, and access control, and presentation attacks are a significant vulnerability of these systems. 
         
         In this paper, we identify the state of the art of presentation attack detection approaches in academia and industry. Then we introduce an end-to-end pipeline for presentation attack detection approaches. Most of the presentation attack detection approaches in the literature detect presentation attacks using machine/deep learning techniques. Due to the various attack types, it is challenging to collect training data sets that include all attack types. In addition, these models are restricted to inputs of a particular minimum quality and are CPU and memory-intensive. Hence, in this study, we proposed a novel, user-friendly CPU-light approach that requires no training phase and is highly robust to 2D attacks. We have tested our service across various 2D attacks, such as digital photos, videos, and printed photos, with users of various ages. We proposed an automated online exam proctoring framework using the identified presentation attack detection approaches as a second contribution to this paper. 
    
    \subsection{Future Research Areas}
        Based on the weak points of the approach proposed in this paper and on analysing the literature and understanding the challenges and future needs of PAD, we offer some potential future research areas to extend this work.
        \\\\
        \textbf{Extend The Proposed Service to Detect 3-Dimensional attacks}. 
            The first future path for this work is extending it to detect 3D presentation attacks in addition to 2D attacks. The fundamentals of the test are to find the face's spatial feature using specific movements; basically, it cannot detect 3D attacks. However, adding some light texture/life sign-based models can also make the system robust to 3D attacks. The challenge here is to develop a light model that works with low-quality data and preferably does not need any training stage to let the model maintain its contributions to solving these gaps. 
        \\\\    
        \textbf{Anomaly Detection and Domain Generalization}
            As technology advances, more realistic and high-resolution attacks are developed. This increases the need to develop PAD approaches in order to detect new attack kinds. Because the majority of PAD approaches involve machine learning algorithms, a training stage is required in which the method learns different attack types. One strategy we can use to stay up to date on new attacks is to use methods that can detect previously unseen attacks. For this reason, anomaly detection and domain generalisation techniques have recently become prominent in PAD. However, these methods require to be more analysed and debated in a variety of ways. 
        \\\\    
        \textbf{Deepfake Attack Detection}
            Deepfake is one of the more recent attacks. With the aid of Deepfake technology, a person's face is swapped for their original match in a picture or video. It creates high-quality and natural fake face videos with publicly available technology. In live capture scenarios (when the frames are lively captured by a camera sensor), this attack will be converted to the replay video attack. However, in API-based approaches (image/video is sent to an API), this technology can cause spoofing. Deepfake attack detection has been the subject of numerous studies. Nevertheless, further research and improvements are still needed.
        \\\\
        \textbf{Develop the Online Exam Proctoring Framework}
            We offered an automated online exam proctoring framework in this paper which, to the best of our knowledge, is first to analyse the heart rate of the exam candidate using presentation attack detection approaches. This approach could be expanded and developed to be used in various institutions, including various test types.


\section*{Acknowledgements}
- I acknowledge the Centre for Applied Artificial Intelligence\footnote{https://www.mq.edu.au/research/research-centres-groups-and-facilities/centres/centre-for-applied-artificial-intelligence} at Macquarie University and Locii Innovation Pty Ltd (trades under the business name trUUth\footnote{https://www.truuth.id/}) for funding My Master by Research project.

\bibliographystyle{abbrv}
\bibliography{main}

\begin{thebibliography}{100}

\bibitem{mediapipe}
Face mesh, \url{https://google.github.io/mediapipe/solutions/face_mesh.html}.

\bibitem{liveness_website}
Liveness.com - biometric liveness detection explained,
  \url{https://www.liveness.com/#free}.

\bibitem{experian_2021}
What is the difference between validation, verification and authentication, Sep
  2021.

\bibitem{ibeta_2022}
Iso 30107-3 presentation attack detection test methodology and confirmation
  letters, Oct 2022.

\bibitem{agarwal2016face}
A.~Agarwal, R.~Singh, and M.~Vatsa.
\newblock Face anti-spoofing using haralick features.
\newblock In {\em 2016 IEEE 8th International Conference on Biometrics Theory,
  Applications and Systems (BTAS)}, pages 1--6. IEEE, 2016.

\bibitem{akbulut2017deep}
Y.~Akbulut, A.~{\c{S}}eng{\"u}r, {\"U}.~Budak, and S.~Ekici.
\newblock Deep learning based face liveness detection in videos.
\newblock In {\em 2017 international artificial intelligence and data
  processing symposium (IDAP)}, pages 1--4. Ieee, 2017.

\bibitem{alahmid_2020}
M.~Alahmid.
\newblock Face recognition system and calculating frr, far and eer for
  biometric system evaluation + code, 2020.

\bibitem{anjos2011counter}
A.~Anjos and S.~Marcel.
\newblock Counter-measures to photo attacks in face recognition: a public
  database and a baseline.
\newblock In {\em 2011 international joint conference on Biometrics (IJCB)},
  pages 1--7. IEEE, 2011.

\bibitem{arab_2022}
O.~Arab.
\newblock Council post: What the rapid adoption of biometrics means for your
  business, Jan 2022.

\bibitem{arashloo2020unseen}
S.~R. Arashloo.
\newblock Unseen face presentation attack detection using sparse multiple
  kernel fisher null-space.
\newblock {\em IEEE Transactions on Circuits and Systems for Video Technology},
  31(10):4084--4095, 2020.

\bibitem{arashloo2017anomaly}
S.~R. Arashloo, J.~Kittler, and W.~Christmas.
\newblock An anomaly detection approach to face spoofing detection: A new
  formulation and evaluation protocol.
\newblock {\em IEEE access}, 5:13868--13882, 2017.

\bibitem{ashraf2020novel}
J.~Ashraf, A.~D. Bakhshi, N.~Moustafa, H.~Khurshid, A.~Javed, and A.~Beheshti.
\newblock Novel deep learning-enabled lstm autoencoder architecture for
  discovering anomalous events from intelligent transportation systems.
\newblock {\em IEEE Transactions on Intelligent Transportation Systems},
  22(7):4507--4518, 2020.

\bibitem{atoum2017automated}
Y.~Atoum, L.~Chen, A.~X. Liu, S.~D. Hsu, and X.~Liu.
\newblock Automated online exam proctoring.
\newblock {\em IEEE Transactions on Multimedia}, 19(7):1609--1624, 2017.

\bibitem{aware_2022}
Aware.
\newblock Knomi: Mobile biometric authentication, 2022.

\bibitem{liveness_detection_framework_1991}
AWS.
\newblock Liveness detection framework, 2022.

\bibitem{baweja2020anomaly}
Y.~Baweja, P.~Oza, P.~Perera, and V.~M. Patel.
\newblock Anomaly detection-based unknown face presentation attack detection.
\newblock In {\em 2020 IEEE International Joint Conference on Biometrics
  (IJCB)}, pages 1--9. IEEE, 2020.

\bibitem{beheshti2017coredb}
A.~Beheshti, B.~Benatallah, R.~Nouri, V.~M. Chhieng, H.~Xiong, and X.~Zhao.
\newblock Coredb: a data lake service.
\newblock In {\em Proceedings of the 2017 ACM on Conference on Information and
  Knowledge Management}, pages 2451--2454, 2017.

\bibitem{beheshti2018corekg}
A.~Beheshti, B.~Benatallah, R.~Nouri, and A.~Tabebordbar.
\newblock Corekg: a knowledge lake service.
\newblock {\em Proceedings of the VLDB Endowment}, 11(12):1942--1945, 2018.

\bibitem{beheshti2020intelligent}
A.~Beheshti, B.~Benatallah, Q.~Z. Sheng, and F.~Schiliro.
\newblock Intelligent knowledge lakes: The age of artificial intelligence and
  big data.
\newblock In {\em International conference on web information systems
  engineering}, pages 24--34. Springer, 2020.

\bibitem{beheshti2019datasynapse}
A.~Beheshti, B.~Benatallah, A.~Tabebordbar, H.~R. Motahari-Nezhad, M.~C.
  Barukh, and R.~Nouri.
\newblock Datasynapse: A social data curation foundry.
\newblock {\em Distributed and Parallel Databases}, 37(3):351--384, 2019.

\bibitem{beheshti2018iprocess}
A.~Beheshti, F.~Schiliro, S.~Ghodratnama, F.~Amouzgar, B.~Benatallah, J.~Yang,
  Q.~Z. Sheng, F.~Casati, and H.~R. Motahari-Nezhad.
\newblock iprocess: Enabling iot platforms in data-driven knowledge-intensive
  processes.
\newblock In {\em International Conference on Business Process Management},
  pages 108--126. Springer, 2018.

\bibitem{beheshti2020towards}
A.~Beheshti, S.~Yakhchi, S.~Mousaeirad, S.~M. Ghafari, S.~R. Goluguri, and
  M.~A. Edrisi.
\newblock Towards cognitive recommender systems.
\newblock {\em Algorithms}, 13(8):176, 2020.

\bibitem{beheshti2016scalable}
S.-M.-R. Beheshti, B.~Benatallah, and H.~R. Motahari-Nezhad.
\newblock Scalable graph-based olap analytics over process execution data.
\newblock {\em Distributed and Parallel Databases}, 34(3):379--423, 2016.

\bibitem{benatallah2016process}
B.~Benatallah, S.~Sakr, D.~Grigori, H.~R. Motahari-Nezhad, M.~C. Barukh,
  A.~Gater, S.~H. Ryu, et~al.
\newblock {\em Process analytics: concepts and techniques for querying and
  analyzing process data}.
\newblock Springer, 2016.

\bibitem{bharadwaj2013computationally}
S.~Bharadwaj, T.~I. Dhamecha, M.~Vatsa, and R.~Singh.
\newblock Computationally efficient face spoofing detection with motion
  magnification.
\newblock In {\em Proceedings of the IEEE conference on computer vision and
  pattern recognition workshops}, pages 105--110, 2013.

\bibitem{administrator_2022}
BioID.
\newblock Face liveness detection: Anti-spoofing: Biometrics: Pad, 2022.

\bibitem{boulkenafet2015face}
Z.~Boulkenafet, J.~Komulainen, and A.~Hadid.
\newblock Face anti-spoofing based on color texture analysis.
\newblock In {\em 2015 IEEE international conference on image processing
  (ICIP)}, pages 2636--2640. IEEE, 2015.

\bibitem{boulkenafet2017oulu}
Z.~Boulkenafet, J.~Komulainen, L.~Li, X.~Feng, and A.~Hadid.
\newblock Oulu-npu: A mobile face presentation attack database with real-world
  variations.
\newblock In {\em 2017 12th IEEE international conference on automatic face \&
  gesture recognition (FG 2017)}, pages 612--618. IEEE, 2017.

\bibitem{bousnina2021unraveling}
N.~Bousnina, L.~Zheng, M.~Mikram, S.~Ghouzali, and K.~Minaoui.
\newblock Unraveling robustness of deep face anti-spoofing models against pixel
  attacks.
\newblock {\em Multimedia Tools and Applications}, 80(5):7229--7246, 2021.

\bibitem{cordennebrewster}
C.~Brewste.
\newblock Applications of facial recognition technology, 2019.

\bibitem{chen2019cascade}
H.~Chen, Y.~Chen, X.~Tian, and R.~Jiang.
\newblock A cascade face spoofing detector based on face anti-spoofing r-cnn
  and improved retinex lbp.
\newblock {\em IEEE Access}, 7:170116--170133, 2019.

\bibitem{chingovska2012effectiveness}
I.~Chingovska, A.~Anjos, and S.~Marcel.
\newblock On the effectiveness of local binary patterns in face anti-spoofing.
\newblock In {\em 2012 BIOSIG-proceedings of the international conference of
  biometrics special interest group (BIOSIG)}, pages 1--7. IEEE, 2012.

\bibitem{chingovska2019evaluation}
I.~Chingovska, A.~Mohammadi, A.~Anjos, and S.~Marcel.
\newblock Evaluation methodologies for biometric presentation attack detection.
\newblock In {\em handbook of biometric anti-spoofing}, pages 457--480.
  Springer, 2019.

\bibitem{costa2016replay}
A.~Costa-Pazo, S.~Bhattacharjee, E.~Vazquez-Fernandez, and S.~Marcel.
\newblock The replay-mobile face presentation-attack database.
\newblock In {\em 2016 International Conference of the Biometrics Special
  Interest Group (BIOSIG)}, pages 1--7. IEEE, 2016.

\bibitem{de2013can}
T.~de~Freitas~Pereira, A.~Anjos, J.~M. De~Martino, and S.~Marcel.
\newblock Can face anti-spoofing countermeasures work in a real world scenario?
\newblock In {\em 2013 international conference on biometrics (ICB)}, pages
  1--8. IEEE, 2013.

\bibitem{de2012moving}
M.~De~Marsico, M.~Nappi, D.~Riccio, and J.-L. Dugelay.
\newblock Moving face spoofing detection via 3d projective invariants.
\newblock In {\em 2012 5th IAPR International Conference on Biometrics (ICB)},
  pages 73--78. IEEE, 2012.

\bibitem{dorazio2022}
J.~Dorazio.
\newblock Council post: Growing an online business: More than having the right
  digital tools, Apr 2022.

\bibitem{ebrary}
Ebrary.
\newblock Feature extraction, Oct 2014.

\bibitem{elahi2021recommender}
M.~Elahi, A.~Beheshti, and S.~R. Goluguri.
\newblock Recommender systems: Challenges and opportunities in the age of big
  data and artificial intelligence.
\newblock {\em Data Science and Its Applications}, pages 15--39, 2021.

\bibitem{elecID}
ElectronicID.
\newblock Smileid: The new standard for face biometrics, 2022.

\bibitem{facetec}
FaceTec.
\newblock 3d liveness, ocr, and 3d face authentication software, 2022.

\bibitem{farhood2022recent}
H.~Farhood, I.~Bakhshayeshi, M.~Pooshideh, N.~Rezvani, and A.~Beheshti.
\newblock Recent advances of image processing techniques in agriculture.
\newblock {\em Artificial Intelligence and Data Science in Environmental
  Sensing}, pages 129--153, 2022.

\bibitem{fatemifar2021client}
S.~Fatemifar, S.~R. Arashloo, M.~Awais, and J.~Kittler.
\newblock Client-specific anomaly detection for face presentation attack
  detection.
\newblock {\em Pattern Recognition}, 112:107696, 2021.

\bibitem{fernandez2016driver}
A.~Fern{\'a}ndez, R.~Usamentiaga, J.~L. Car{\'u}s, and R.~Casado.
\newblock Driver distraction using visual-based sensors and algorithms.
\newblock {\em Sensors}, 16(11):1805, 2016.

\bibitem{fourati2020anti}
E.~Fourati, W.~Elloumi, and A.~Chetouani.
\newblock Anti-spoofing in face recognition-based biometric authentication
  using image quality assessment.
\newblock {\em Multimedia Tools and Applications}, 79(1):865--889, 2020.

\bibitem{freitas2012lbp}
T.~d. Freitas~Pereira, A.~Anjos, J.~M.~D. Martino, and S.~Marcel.
\newblock Lbp- top based countermeasure against face spoofing attacks.
\newblock In {\em Asian Conference on Computer Vision}, pages 121--132.
  Springer, 2012.

\bibitem{freitas2014face}
T.~d. Freitas~Pereira, J.~Komulainen, A.~Anjos, J.~M. De~Martino, A.~Hadid,
  M.~Pietik{\"a}inen, and S.~Marcel.
\newblock Face liveness detection using dynamic texture.
\newblock {\em EURASIP Journal on Image and video Processing}, 2014(1):1--15,
  2014.

\bibitem{galbally2013image}
J.~Galbally, S.~Marcel, and J.~Fierrez.
\newblock Image quality assessment for fake biometric detection: Application to
  iris, fingerprint, and face recognition.
\newblock {\em IEEE transactions on image processing}, 23(2):710--724, 2013.

\bibitem{george2020learning}
A.~George and S.~Marcel.
\newblock Learning one class representations for face presentation attack
  detection using multi-channel convolutional neural networks.
\newblock {\em IEEE Transactions on Information Forensics and Security},
  16:361--375, 2020.

\bibitem{ghodratnama2021summary2vec}
S.~Ghodratnama, M.~Zakershahrak, and A.~Beheshti.
\newblock Summary2vec: Learning semantic representation of summaries for
  healthcare analytics.
\newblock In {\em 2021 International Joint Conference on Neural Networks
  (IJCNN)}, pages 1--8. IEEE, 2021.

\bibitem{guennouni2019biometric}
S.~Guennouni, A.~Mansouri, and A.~Ahaitouf.
\newblock Biometric systems and their applications.
\newblock In {\em Visual Impairment and Blindness-What We Know and What We Have
  to Know}. IntechOpen, 2019.

\bibitem{hernandez2021biometric}
M.~Hernandez-de Menendez, R.~Morales-Menendez, C.~A. Escobar, and J.~Arinez.
\newblock Biometric applications in education.
\newblock {\em International Journal on Interactive Design and Manufacturing
  (IJIDeM)}, 15(2):365--380, 2021.

\bibitem{hussein2020evaluation}
M.~J. Hussein, J.~Yusuf, A.~S. Deb, L.~Fong, and S.~Naidu.
\newblock An evaluation of online proctoring tools.
\newblock {\em Open Praxis}, 12(4):509--525, 2020.

\bibitem{idcentral}
IDcentral.
\newblock What is liveness detection? types and benefits of liveness detection,
  2022.

\bibitem{Idrnd}
Idrnd.
\newblock Better biometric authentication starts with idrnd, 2022.

\bibitem{idrus2013review}
S.~Z.~S. Idrus, E.~Cherrier, C.~Rosenberger, and J.-J. Schwartzmann.
\newblock A review on authentication methods.
\newblock {\em Australian Journal of Basic and Applied Sciences}, 7(5):95--107,
  2013.

\bibitem{iso_2021}
ISO.
\newblock Iso/iec 30107-1:2016, 2016.

\bibitem{jee2006liveness}
H.-K. Jee, S.-U. Jung, and J.-H. Yoo.
\newblock Liveness detection for embedded face recognition system.
\newblock {\em International Journal of Biological and Medical Sciences},
  1(4):235--238, 2006.

\bibitem{jia2019database}
S.~Jia, C.~Hu, G.~Guo, and Z.~Xu.
\newblock A database for face presentation attack using wax figure faces.
\newblock In {\em International Conference on Image Analysis and Processing},
  pages 39--47. Springer, 2019.

\bibitem{jia2021face}
S.~Jia, C.~Hu, X.~Li, and Z.~Xu.
\newblock Face spoofing detection under super-realistic 3d wax face attacks.
\newblock {\em Pattern Recognition Letters}, 145:103--109, 2021.

\bibitem{jia20203d}
S.~Jia, X.~Li, C.~Hu, G.~Guo, and Z.~Xu.
\newblock 3d face anti-spoofing with factorized bilinear coding.
\newblock {\em IEEE Transactions on Circuits and Systems for Video Technology},
  31(10):4031--4045, 2020.

\bibitem{kasinathan2022proctorex}
V.~Kasinathan, C.~E. Yan, A.~Mustapha, V.~A. Hameed, T.~H. Ching, and
  V.~Thiruchelvam.
\newblock Proctorex: An automated online exam proctoring system.
\newblock {\em Mathematical Statistician and Engineering Applications},
  71(3s2):876--889, 2022.

\bibitem{khatami2020convolutional}
A.~Khatami, A.~Nazari, A.~Beheshti, T.~T. Nguyen, S.~Nahavandi, and J.~Zieba.
\newblock Convolutional neural network for medical image classification using
  wavelet features.
\newblock In {\em 2020 International Joint Conference on Neural Networks
  (IJCNN)}, pages 1--8. IEEE, 2020.

\bibitem{kim2011motion}
Y.~Kim, J.-H. Yoo, and K.~Choi.
\newblock A motion and similarity-based fake detection method for biometric
  face recognition systems.
\newblock {\em IEEE Transactions on Consumer Electronics}, 57(2):756--762,
  2011.

\bibitem{kollreider2009non}
K.~Kollreider, H.~Fronthaler, and J.~Bigun.
\newblock Non-intrusive liveness detection by face images.
\newblock {\em Image and Vision Computing}, 27(3):233--244, 2009.

\bibitem{komulainen2013complementary}
J.~Komulainen, A.~Hadid, M.~Pietik{\"a}inen, A.~Anjos, and S.~Marcel.
\newblock Complementary countermeasures for detecting scenic face spoofing
  attacks.
\newblock In {\em 2013 International conference on biometrics (ICB)}, pages
  1--7. IEEE, 2013.

\bibitem{koppikar2021face}
U.~Koppikar, C.~Sujatha, P.~Patil, and P.~Hiremath.
\newblock Face liveness detection to overcome spoofing attacks in face
  recognition system.
\newblock In {\em Innovations in Computational Intelligence and Computer
  Vision}, pages 351--360. Springer, 2021.

\bibitem{kose2013countermeasure}
N.~Kose and J.-L. Dugelay.
\newblock Countermeasure for the protection of face recognition systems against
  mask attacks.
\newblock In {\em 2013 10th IEEE International Conference and Workshops on
  Automatic Face and Gesture Recognition (FG)}, pages 1--6. IEEE, 2013.

\bibitem{koshy2019optimizing}
R.~Koshy and A.~Mahmood.
\newblock Optimizing deep cnn architectures for face liveness detection.
\newblock {\em Entropy}, 21(4):423, 2019.

\bibitem{lavrentyeva2018interactive}
G.~Lavrentyeva, O.~Kudashev, A.~Melnikov, M.~D. Marsico, and Y.~Matveev.
\newblock Interactive photo liveness for presentation attacks detection.
\newblock In {\em International Conference Image Analysis and Recognition},
  pages 252--258. Springer, 2018.

\bibitem{li2018unsupervised}
H.~Li, W.~Li, H.~Cao, S.~Wang, F.~Huang, and A.~C. Kot.
\newblock Unsupervised domain adaptation for face anti-spoofing.
\newblock {\em IEEE Transactions on Information Forensics and Security},
  13(7):1794--1809, 2018.

\bibitem{li2004live}
J.~Li, Y.~Wang, T.~Tan, and A.~K. Jain.
\newblock Live face detection based on the analysis of fourier spectra.
\newblock In {\em Biometric technology for human identification}, volume 5404,
  pages 296--303. SPIE, 2004.

\bibitem{li2020compactnet}
L.~Li, Z.~Xia, X.~Jiang, F.~Roli, and X.~Feng.
\newblock Compactnet: learning a compact space for face presentation attack
  detection.
\newblock {\em neurocomputing}, 409:191--207, 2020.

\bibitem{li2016generalized}
X.~Li, J.~Komulainen, G.~Zhao, P.-C. Yuen, and M.~Pietik{\"a}inen.
\newblock Generalized face anti-spoofing by detecting pulse from face videos.
\newblock In {\em 2016 23rd International Conference on Pattern Recognition
  (ICPR)}, pages 4244--4249. IEEE, 2016.

\bibitem{li2016face}
Y.~Li, L.-M. Po, X.~Xu, L.~Feng, and F.~Yuan.
\newblock Face liveness detection and recognition using shearlet based feature
  descriptors.
\newblock In {\em 2016 IEEE International Conference on Acoustics, Speech and
  Signal Processing (ICASSP)}, pages 874--877. IEEE, 2016.

\bibitem{lim2020one}
S.~Lim, Y.~Gwak, W.~Kim, J.-H. Roh, and S.~Cho.
\newblock One-class learning method based on live correlation loss for face
  anti-spoofing.
\newblock {\em IEEE Access}, 8:201635--201648, 2020.

\bibitem{lin2019face}
B.~Lin, X.~Li, Z.~Yu, and G.~Zhao.
\newblock Face liveness detection by rppg features and contextual patch-based
  cnn.
\newblock In {\em Proceedings of the 2019 3rd international conference on
  biometric engineering and applications}, pages 61--68, 2019.

\bibitem{lin2019convolutional}
H.-Y.~S. Lin and Y.-W. Su.
\newblock Convolutional neural networks for face anti-spoofing and liveness
  detection.
\newblock In {\em 2019 6th International Conference on Systems and Informatics
  (ICSAI)}, pages 1233--1237. IEEE, 2019.

\bibitem{liu20163dd}
S.~Liu, B.~Yang, P.~C. Yuen, and G.~Zhao.
\newblock A 3d mask face anti-spoofing database with real world variations.
\newblock In {\em Proceedings of the IEEE conference on computer vision and
  pattern recognition workshops}, pages 100--106, 2016.

\bibitem{liu20163d}
S.~Liu, P.~C. Yuen, S.~Zhang, and G.~Zhao.
\newblock 3d mask face anti-spoofing with remote photoplethysmography.
\newblock In {\em European Conference on Computer Vision}, pages 85--100.
  Springer, 2016.

\bibitem{liu2018learning}
Y.~Liu, A.~Jourabloo, and X.~Liu.
\newblock Learning deep models for face anti-spoofing: Binary or auxiliary
  supervision.
\newblock In {\em Proceedings of the IEEE conference on computer vision and
  pattern recognition}, pages 389--398, 2018.

\bibitem{liu2019presentation}
Y.~Liu, J.~Stehouwer, A.~Jourabloo, Y.~Atoum, and X.~Liu.
\newblock Presentation attack detection for face in mobile phones.
\newblock In {\em Selfie Biometrics}, pages 171--196. Springer, 2019.

\bibitem{Luxand}
Luxand.
\newblock Luxand facesdk – liveness detection, 2022.

\bibitem{ma2020novel}
Y.~Ma, L.~Wu, Z.~Li, et~al.
\newblock A novel face presentation attack detection scheme based on
  multi-regional convolutional neural networks.
\newblock {\em Pattern Recognition Letters}, 131:261--267, 2020.

\bibitem{ma2020multi}
Y.~Ma, Y.~Xu, and F.~Liu.
\newblock Multi-perspective dynamic features for cross-database face
  presentation attack detection.
\newblock {\em IEEE Access}, 8:26505--26516, 2020.

\bibitem{maatta2011face}
J.~M{\"a}{\"a}tt{\"a}, A.~Hadid, and M.~Pietik{\"a}inen.
\newblock Face spoofing detection from single images using micro-texture
  analysis.
\newblock In {\em 2011 international joint conference on Biometrics (IJCB)},
  pages 1--7. IEEE, 2011.

\bibitem{maatta2012face}
J.~M{\"a}{\"a}tt{\"a}, A.~Hadid, and M.~Pietik{\"a}inen.
\newblock Face spoofing detection from single images using texture and local
  shape analysis.
\newblock {\em IET biometrics}, 1(1):3--10, 2012.

\bibitem{mu2019face}
D.~Mu and T.~Li.
\newblock Face anti-spoofing with multi-color double-stream cnn.
\newblock In {\em Proceedings of the 13th International Conference on
  Distributed Smart Cameras}, pages 1--4, 2019.

\bibitem{gavin_2021}
NEC.
\newblock The top 9 common uses of biometrics in everyday life - nec nz, 2021.

\bibitem{nesli2013spoofing}
E.~Nesli and S.~Marcel.
\newblock Spoofing in 2d face recognition with 3d masks and anti-spoofing with
  kinect.
\newblock In {\em IEEE 6th International Conference on Biometrics: Theory,
  Applications and Systems (BTAS’13)}, pages 1--8, 2013.

\bibitem{nguyen2019face}
H.~P. Nguyen, A.~Delahaies, F.~Retraint, and F.~Morain-Nicolier.
\newblock Face presentation attack detection based on a statistical model of
  image noise.
\newblock {\em IEEE Access}, 7:175429--175442, 2019.

\bibitem{nowara2017ppgsecure}
E.~M. Nowara, A.~Sabharwal, and A.~Veeraraghavan.
\newblock Ppgsecure: Biometric presentation attack detection using
  photopletysmograms.
\newblock In {\em 2017 12th IEEE International Conference on Automatic Face \&
  Gesture Recognition (FG 2017)}, pages 56--62. IEEE, 2017.

\bibitem{pan2007eyeblink}
G.~Pan, L.~Sun, Z.~Wu, and S.~Lao.
\newblock Eyeblink-based anti-spoofing in face recognition from a generic
  webcamera.
\newblock In {\em 2007 IEEE 11th international conference on computer vision},
  pages 1--8. IEEE, 2007.

\bibitem{patel2016cross}
K.~Patel, H.~Han, and A.~K. Jain.
\newblock Cross-database face antispoofing with robust feature representation.
\newblock In {\em Chinese Conference on Biometric Recognition}, pages 611--619.
  Springer, 2016.

\bibitem{patel2016secure}
K.~Patel, H.~Han, and A.~K. Jain.
\newblock Secure face unlock: Spoof detection on smartphones.
\newblock {\em IEEE transactions on information forensics and security},
  11(10):2268--2283, 2016.

\bibitem{patel2015live}
K.~Patel, H.~Han, A.~K. Jain, and G.~Ott.
\newblock Live face video vs. spoof face video: Use of moir{\'e} patterns to
  detect replay video attacks.
\newblock In {\em 2015 International Conference on Biometrics (ICB)}, pages
  98--105. IEEE, 2015.

\bibitem{peixoto2011face}
B.~Peixoto, C.~Michelassi, and A.~Rocha.
\newblock Face liveness detection under bad illumination conditions.
\newblock In {\em 2011 18th IEEE International Conference on Image Processing},
  pages 3557--3560. IEEE, 2011.

\bibitem{peng2018face}
F.~Peng, L.~Qin, and M.~Long.
\newblock Face presentation attack detection using guided scale texture.
\newblock {\em Multimedia Tools and Applications}, 77(7):8883--8909, 2018.

\bibitem{peng2020face}
F.~Peng, L.~Qin, and M.~Long.
\newblock Face presentation attack detection based on chromatic co-occurrence
  of local binary pattern and ensemble learning.
\newblock {\em Journal of Visual Communication and Image Representation},
  66:102746, 2020.

\bibitem{perez2019deep}
D.~P{\'e}rez-Cabo, D.~Jim{\'e}nez-Cabello, A.~Costa-Pazo, and R.~J.
  L{\'o}pez-Sastre.
\newblock Deep anomaly detection for generalized face anti-spoofing.
\newblock In {\em Proceedings of the IEEE/CVF Conference on Computer Vision and
  Pattern Recognition Workshops}, pages 0--0, 2019.

\bibitem{pinto2015face}
A.~Pinto, H.~Pedrini, W.~R. Schwartz, and A.~Rocha.
\newblock Face spoofing detection through visual codebooks of spectral temporal
  cubes.
\newblock {\em IEEE Transactions on Image Processing}, 24(12):4726--4740, 2015.

\bibitem{raghavendra2017extended}
R.~Raghavendra, K.~B. Raja, S.~Venkatesh, and C.~Busch.
\newblock Extended multispectral face presentation attack detection: An
  approach based on fusing information from individual spectral bands.
\newblock In {\em 2017 20th International Conference on Information Fusion
  (Fusion)}, pages 1--6. IEEE, 2017.

\bibitem{ramachandra2017presentation}
R.~Ramachandra and C.~Busch.
\newblock Presentation attack detection methods for face recognition systems: A
  comprehensive survey.
\newblock {\em ACM Computing Surveys (CSUR)}, 50(1):1--37, 2017.

\bibitem{rathgeb2020makeup}
C.~Rathgeb, P.~Drozdowski, and C.~Busch.
\newblock Makeup presentation attacks: Review and detection performance
  benchmark.
\newblock {\em IEEE Access}, 8:224958--224973, 2020.

\bibitem{rehman2020enhancing}
Y.~A.~U. Rehman, L.-M. Po, and J.~Komulainen.
\newblock Enhancing deep discriminative feature maps via perturbation for face
  presentation attack detection.
\newblock {\em Image and Vision Computing}, 94:103858, 2020.

\bibitem{rehman2019perturbing}
Y.~A.~U. Rehman, L.-M. Po, M.~Liu, Z.~Zou, and W.~Ou.
\newblock Perturbing convolutional feature maps with histogram of oriented
  gradients for face liveness detection.
\newblock In {\em International Joint Conference: 12th International Conference
  on Computational Intelligence in Security for Information Systems (CISIS
  2019) and 10th International Conference on EUropean Transnational Education
  (ICEUTE 2019)}, pages 3--13. Springer, 2019.

\bibitem{rehman2019face}
Y.~A.~U. Rehman, L.-M. Po, M.~Liu, Z.~Zou, W.~Ou, and Y.~Zhao.
\newblock Face liveness detection using convolutional-features fusion of real
  and deep network generated face images.
\newblock {\em Journal of Visual Communication and Image Representation},
  59:574--582, 2019.

\bibitem{sankey2021covid}
M.~Sankey.
\newblock Covid-19 exam software survey 2020.
\newblock {\em An ACODE Whitepaper https://www. acode. edu. au/pluginfile.
  php/8244/mod\_resource/content/2/eExamsWhitepaper. pdf. Accessed}, 18, 2021.

\bibitem{schiliro2018icop}
F.~Schiliro, A.~Beheshti, S.~Ghodratnama, F.~Amouzgar, B.~Benatallah, J.~Yang,
  Q.~Z. Sheng, F.~Casati, and H.~R. Motahari-Nezhad.
\newblock icop: Iot-enabled policing processes.
\newblock In {\em International Conference on Service-Oriented Computing},
  pages 447--452. Springer, 2018.

\bibitem{schiliro2020novel}
F.~Schiliro, A.~Beheshti, and N.~Moustafa.
\newblock A novel cognitive computing technique using convolutional networks
  for automating the criminal investigation process in policing.
\newblock In {\em Proceedings of SAI Intelligent Systems Conference}, pages
  528--539. Springer, 2020.

\bibitem{schiliro2020cognitive}
F.~Schiliro, N.~Moustafa, and A.~Beheshti.
\newblock Cognitive privacy: Ai-enabled privacy using eeg signals in the
  internet of things.
\newblock In {\em 2020 IEEE 6th International Conference on Dependability in
  Sensor, Cloud and Big Data Systems and Application (DependSys)}, pages
  73--79. IEEE, 2020.

\bibitem{selwyn2021necessary}
N.~Selwyn, C.~O’Neill, G.~Smith, M.~Andrejevic, and X.~Gu.
\newblock A necessary evil? the rise of online exam proctoring in australian
  universities.
\newblock {\em Media International Australia}, page 1329878X211005862, 2021.

\bibitem{sepas2018light}
A.~Sepas-Moghaddam, F.~Pereira, and P.~L. Correia.
\newblock Light field-based face presentation attack detection: reviewing,
  benchmarking and one step further.
\newblock {\em IEEE Transactions on Information Forensics and Security},
  13(7):1696--1709, 2018.

\bibitem{shabani2022icreate}
N.~Shabani, A.~Beheshti, H.~Farhood, M.~Bower, M.~Garrett, and H.~A. Rokny.
\newblock icreate: Mining creative thinking patterns from contextualized
  educational data.
\newblock In {\em International Conference on Artificial Intelligence in
  Education}, pages 352--356. Springer, 2022.

\bibitem{smiatacz2012liveness}
M.~Smiatacz.
\newblock Liveness measurements using optical flow for biometric person
  authentication.
\newblock {\em Metrology and Measurement Systems}, 19(2):257--268, 2012.

\bibitem{SpoofSence}
SpoofSense.
\newblock Spoofsense - face liveness detection, 2022.

\bibitem{standard2017information}
I.~Standard.
\newblock Information technology--biometric presentation attack detection--part
  3: testing and reporting.
\newblock {\em International Organization for Standardization: Geneva,
  Switzerland}, 2017.

\bibitem{strohm_2022}
M.~Strohm.
\newblock Digital banking survey: How americans prefer to bank, Mar 2022.

\bibitem{Sumsub}
sumsub.
\newblock Quick and secure face authentication with prooface, 2022.

\bibitem{sun2020face}
W.~Sun, Y.~Song, H.~Zhao, and Z.~Jin.
\newblock A face spoofing detection method based on domain adaptation and
  lossless size adaptation.
\newblock {\em IEEE access}, 8:66553--66563, 2020.

\bibitem{tan2010face}
X.~Tan, Y.~Li, J.~Liu, and L.~Jiang.
\newblock Face liveness detection from a single image with sparse low rank
  bilinear discriminative model.
\newblock In {\em European Conference on Computer Vision}, pages 504--517.
  Springer, 2010.

\bibitem{thalesgroup}
thalesgroup.
\newblock Biometrics: Definition, use cases, latest news, 2022.

\bibitem{tronci2011fusion}
R.~Tronci, D.~Muntoni, G.~Fadda, M.~Pili, N.~Sirena, G.~Murgia, M.~Ristori,
  S.~Ricerche, and F.~Roli.
\newblock Fusion of multiple clues for photo-attack detection in face
  recognition systems.
\newblock In {\em 2011 International joint conference on biometrics (IJCB)},
  pages 1--6. IEEE, 2011.

\bibitem{tu2020learning}
X.~Tu, Z.~Ma, J.~Zhao, G.~Du, M.~Xie, and J.~Feng.
\newblock Learning generalizable and identity-discriminative representations
  for face anti-spoofing.
\newblock {\em ACM Transactions on Intelligent Systems and Technology (TIST)},
  11(5):1--19, 2020.

\bibitem{yalefacedatabase}
UCDS.
\newblock Yale face database, May 2001.

\bibitem{van2019face}
D.~T. van~der Haar.
\newblock Face antispoofing using shearlets: an empirical study.
\newblock {\em SAIEE Africa Research Journal}, 110(2):94--103, 2019.

\bibitem{viola2001}
P.~Viola and M.~Jones.
\newblock Rapid object detection using a boosted cascade of simple features.
\newblock In {\em Proceedings of the 2001 IEEE computer society conference on
  computer vision and pattern recognition. CVPR 2001}, volume~1, pages I--I.
  Ieee, 2001.

\bibitem{wang2021assessment2vec}
S.~Wang, A.~Beheshti, Y.~Wang, J.~Lu, Q.~Z. Sheng, S.~Elbourn,
  H.~Alinejad-Rokny, and E.~Galanis.
\newblock Assessment2vec: Learning distributed representations of assessments
  to reduce marking workload.
\newblock In {\em International Conference on Artificial Intelligence in
  Education}, pages 384--389. Springer, 2021.

\bibitem{wang2017face}
S.-Y. Wang, S.-H. Yang, Y.-P. Chen, and J.-W. Huang.
\newblock Face liveness detection based on skin blood flow analysis.
\newblock {\em symmetry}, 9(12):305, 2017.

\bibitem{wen2015face}
D.~Wen, H.~Han, and A.~K. Jain.
\newblock Face spoof detection with image distortion analysis.
\newblock {\em IEEE Transactions on Information Forensics and Security},
  10(4):746--761, 2015.

\bibitem{wu2019face}
Y.~Wu and Q.~Zhao.
\newblock Face presentation attack detection based on exclusivity regularized
  attention maps.
\newblock In {\em Proceedings of the 2019 5th International Conference on
  Computing and Artificial Intelligence}, pages 117--122, 2019.

\bibitem{yan2012face}
J.~Yan, Z.~Zhang, Z.~Lei, D.~Yi, and S.~Z. Li.
\newblock Face liveness detection by exploring multiple scenic clues.
\newblock In {\em 2012 12th International Conference on Control Automation
  Robotics \& Vision (ICARCV)}, pages 188--193. IEEE, 2012.

\bibitem{yang2014learn}
J.~Yang, Z.~Lei, and S.~Z. Li.
\newblock Learn convolutional neural network for face anti-spoofing.
\newblock {\em arXiv preprint arXiv:1408.5601}, 2014.

\bibitem{yu2019diffusion}
C.~Yu, C.~Yao, M.~Pei, and Y.~Jia.
\newblock Diffusion-based kernel matrix model for face liveness detection.
\newblock {\em Image and Vision Computing}, 89:88--94, 2019.

\bibitem{zhang2020face}
K.-Y. Zhang, T.~Yao, J.~Zhang, Y.~Tai, S.~Ding, J.~Li, F.~Huang, H.~Song, and
  L.~Ma.
\newblock Face anti-spoofing via disentangled representation learning.
\newblock In {\em European Conference on Computer Vision}, pages 641--657.
  Springer, 2020.

\bibitem{zhang2020casia}
S.~Zhang, A.~Liu, J.~Wan, Y.~Liang, G.~Guo, S.~Escalera, H.~J. Escalante, and
  S.~Z. Li.
\newblock Casia-surf: A large-scale multi-modal benchmark for face
  anti-spoofing.
\newblock {\em IEEE Transactions on Biometrics, Behavior, and Identity
  Science}, 2(2):182--193, 2020.

\bibitem{zhang2012face}
Z.~Zhang, J.~Yan, S.~Liu, Z.~Lei, D.~Yi, and S.~Z. Li.
\newblock A face antispoofing database with diverse attacks.
\newblock In {\em 2012 5th IAPR international conference on Biometrics (ICB)},
  pages 26--31. IEEE, 2012.

\bibitem{zhou2020domain}
J.~Zhou, K.~Shu, D.~Zhao, and Z.~Xia.
\newblock Domain adaptation based person-specific face anti-spoofing using
  color texture features.
\newblock In {\em Proceedings of the 2020 5th International Conference on
  Machine Learning Technologies}, pages 79--85, 2020.

\end{thebibliography}

\end{document}